\documentclass[twoside,11pt]{article}

\usepackage{jmlr2e}
\usepackage[ruled, vlined, linesnumbered]{algorithm2e}
\usepackage[noend]{algpseudocode}
\usepackage{amsmath}
\usepackage{subcaption}

\jmlrheading{1}{}{}{}{}{}

\firstpageno{1}

\begin{document}

\title{Compute-Efficient Active Learning }

\author{\name Gábor Németh \email gabor.nemeth@aimotive.com \\
       \addr aiMotive \\
       Budapest, Hungary
       \AND
       \name Tamás Matuszka \email tamas.matuszka@aimotive.com \\
       \addr aiMotive \\
       Budapest, Hungary}

%\editor{Leslie Pack Kaelbling}

\maketitle

\begin{abstract}%   <- trailing '%' for backward compatibility of .sty file
%Ezt még átirni
Active learning, a powerful paradigm in machine learning, aims at reducing labeling costs by selecting the most informative samples from an unlabeled dataset. However, the traditional active learning process often demands extensive computational resources, hindering scalability and efficiency. In this paper, we address this critical issue by presenting a novel method designed to alleviate the computational burden associated with active learning on massive datasets. To achieve this goal, we introduce a simple, yet effective method-agnostic framework that outlines how to strategically choose and annotate data points, optimizing the process for efficiency while maintaining model performance. Through case studies, we demonstrate the effectiveness of our proposed method in reducing computational costs while maintaining or, in some cases, even surpassing baseline model outcomes. Code is available at \url{https://github.com/aimotive/Compute-Efficient-Active-Learning}.

\end{abstract}

\begin{keywords}
  Active Learning, Large Datasets, Compute-Efficient 
\end{keywords}

\section{Introduction}
Acquiring large amounts of precisely annotated data for training deep neural networks is extremely expensive. Active learning represents one of the alternative solutions to address this challenge. However, when facing large amounts of unlabeled data, active learning can be computationally expensive. Active learning methods can be grouped into two main categories: acquisition function-based or diversity sampling-based methods (\cite{al_survey}). The goal of the acquisition function is to rank each unlabeled sample based on its importance. With the help of this function, data points can be mapped to a value that indicates the expected contribution to the effect on future model training.

The main motivation of our framework is based on the hypothesis that historical values of the acquisition function are good predictors of their future values. This idea is quite intuitive. For example, once a model is certain about its predictions on a given sample, this fact is very unlikely to change. This can be explained by the randomness in the training, especially when using small acquisition sizes.
%While most methods do not curate the pool of unlabeled data, our method samples each iteration from the unlabeled pool to a candidate set of samples.

\section{Related Work}
In the realm of active learning research, it is noteworthy to acknowledge the existence of papers that put forward computationally efficient solutions. While active learning has traditionally aimed at surpassing state-of-the-art performance on benchmark datasets, a growing body of work has shifted its focus toward addressing the practical computational constraints often faced in real-world applications.

Many of the existing works in this domain concentrate on the computation of epistemic uncertainty, typically achieved through Bayesian approximation using techniques such as model ensembling (as exemplified in \cite{al_ensemble}) or Monte Carlo Dropout (as illustrated in \cite{mc_dropout}). Although these methods offer a rigorous mathematical framework for approximating the posterior distribution, they necessitate multiple forward passes, which can be computationally expensive, particularly when applied to large-scale datasets or resource-constrained environments. Each forward pass incurs a substantial computational overhead, significantly impacting the scalability and efficiency of active learning procedures. 

Recent advances in active learning methodologies encompass a range of novel approaches that have demonstrated promise in enhancing the efficiency and effectiveness of data acquisition. For instance, \cite{al_vat} introduced the Virtual Adversarial Active Learning (VirAAL) method, which simplifies the acquisition function by employing the local distribution roughness (LDR) metric derived from virtual adversarial examples. Similarly, \cite{pt4al} adopts a distinctive strategy, utilizing a separate network for pretext task training and selecting the most challenging samples based on high error rates. \cite{al_vaal} presented Variational Adversarial Active Learning (VAAL), a method that leverages adversarial training to align the feature distributions of labeled and unlabeled samples, ranking them according to the discriminator network's output. Another noteworthy approach is the learning loss method by \cite{learning_loss}, which extends the encoder network with an additional head for predicting the loss function of an image, facilitating acquisition function evaluation in a single forward pass. Building on these concepts, \cite{park2023active} introduced a Model Evidence Head (MEH) and a Hierarchical Uncertainty Aggregation (HUA) framework within the evidential deep learning (EDL) paradigm, enabling the calculation of image informativeness for object detection. Notably, HUA takes into account bounding box attributes, departing from the conventional mean/maximum-based approaches. This method's success in training object detection models aligns with the objectives of our proposed approach.

While many of these approaches indeed demonstrate reduced computational requirements compared to uncertainty-based methods, there remains a computational burden when dealing with extensive datasets, necessitating the evaluation of the acquisition function for each unlabeled data point. In contrast, our proposed framework offers a complementary solution to these existing methods, effectively mitigating the computational demands to a considerable extent.

\section{Compute-Efficient Active Learning}
In our methodology, we implement a discerning subsampling technique rooted in historical acquisition function evaluations. At each iteration of our algorithm, we selectively draw a predetermined quantity of unlabeled data points, leveraging their past acquisition function values as guidance. This selection process is thoughtfully designed to assign higher probabilities of inclusion to samples of greater significance, ensuring that the candidate pool is enriched with influential instances.

\begin{algorithm}[H]
\SetKwInOut{Input}{Input}
\SetKwInOut{Output}{Output}
\SetAlgoLined

\Input{Initial labeled dataset $D_{labeled}$, Unlabeled dataset $D_{unlabeled}$, Model $M$, Number of iterations $T$, Subsample ratio $\alpha$}
\Output{Annotated dataset $D_{labeled}$}

$AcquisitionValues \leftarrow \text{AcquisitionFunction}(M, D_{unlabeled})$ \tcp*{Evaluate acquisition function on all unlabeled samples} 

\For{$t \leftarrow 1$ \KwTo $T$}{

    $M \leftarrow \text{TrainModel}(D_{labeled})$
    
    $P \leftarrow \text{softmax}(AcquisitionValues)$

    $N \leftarrow |D_{unlabeled}|$
    
    $CandidatePool \leftarrow \text{Sample}(P, D_{unlabeled})$ with size $\alpha \cdot N$

    $AcquisitionValues[CandidatePool] \leftarrow \text{AcquisitionFunction}(M, CandidatePool)$
    
    $X_{label} \leftarrow \text{TopK}(CandidatePool, AcquisitionValues[CandidatePool])$
    
    $D_{labeled} \leftarrow D_{labeled} \cup X_{labeled}$
    
    $D_{unlabeled} \leftarrow D_{unlabeled} \setminus X_{label}$
}
\caption{Compute-Efficient Active Learning with Subsampling}
\end{algorithm}

Subsequently, our approach shifts its focus exclusively to the candidate pool for the evaluation of the acquisition function and the strategic determination of which samples merit annotation. This multi-step strategy effectively optimizes the allocation of labeling resources by prioritizing samples with greater potential to enhance the model's performance, thereby enhancing the efficiency and efficacy of our active learning framework.

The subsampling strategy employed in our approach is guided by criteria that maximize the informativeness of the selected data points. These criteria are tailored to the specific learning task and can include measures of uncertainty, diversity, or other domain-specific factors. The proposed method is general and can be combined with various acquisition functions. We have conducted experiments (see detailed results in Section \ref{sec:exps}) using Shannon entropy with MC Dropout and Variation Ratios (varR). However, other acquisition functions can also be used, such as ensemble score (\cite{al_ensemble}) or BALD (\cite{al_bald}). Our method is also suitable for regression problems besides classification. One realization for applying the proposed method to regression might be to measure the deviation of the output distribution and use it as the acquisition function value. Our method streamlines the annotation process by focusing on a reduced candidate dataset, decreasing the computational resources required for active learning.  In this way, compute-efficient active learning can be utilized for various problems.

Additionally, we could also incorporate the process of excluding samples from the unlabeled dataset with exceptionally low acquisition function values. Alternatively, it is possible to remove these samples completely. This approach yields a dual benefit: it further diminishes computational requirements while simultaneously improving the approximation of the original acquisition function-based sampling. Consequently, this method contributes to reductions in both computational and storage demands.

\subsection{Experiments}
\label{sec:exps}

\begin{figure}[t]
\begin{subfigure}{.5\textwidth}
  \centering
  \includegraphics[width=1.05\linewidth]{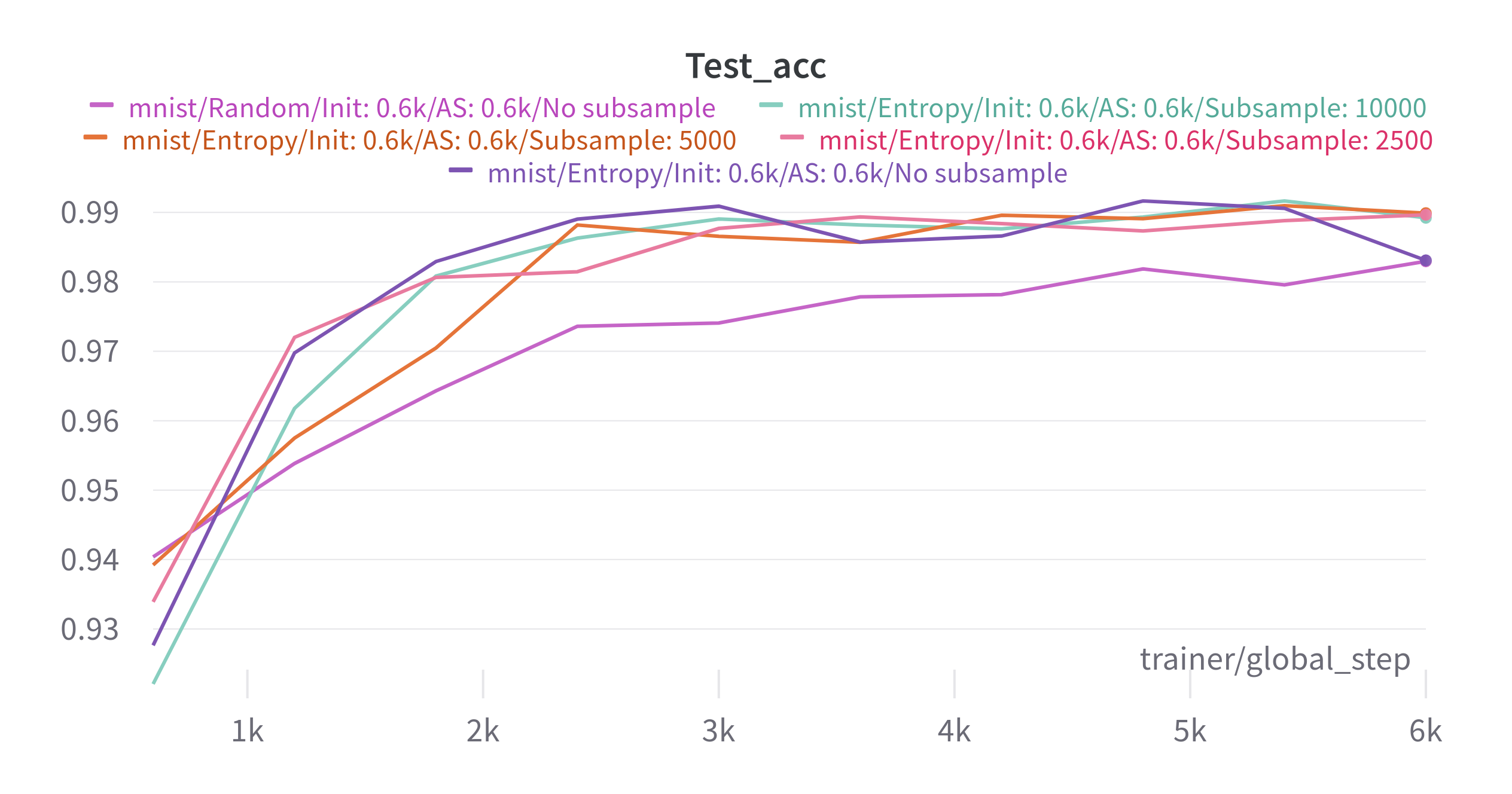}
  \caption{AF: entropy, IP: 1\%, TAS: 10\%}
  \label{fig:mnist_1_10}
\end{subfigure}%
\begin{subfigure}{.5\textwidth}
  \centering
  \includegraphics[width=1.05\linewidth]{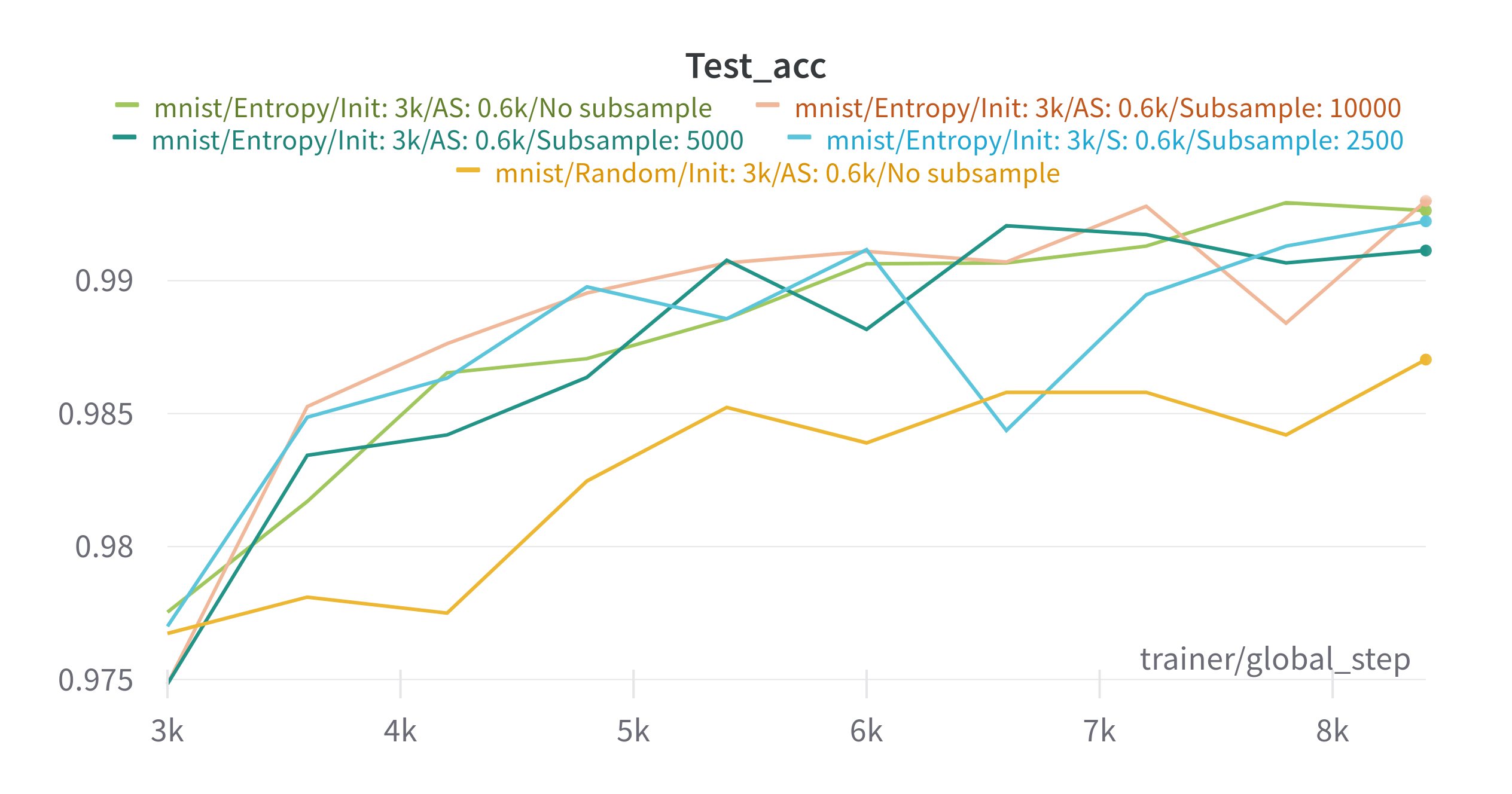}
  \caption{AF: entropy, IP: 5\%, TAS: 10\%}
  \label{fig:mnist_5_10}
\end{subfigure}
\begin{subfigure}{.5\textwidth}
  \centering
  \includegraphics[width=1.05\linewidth]{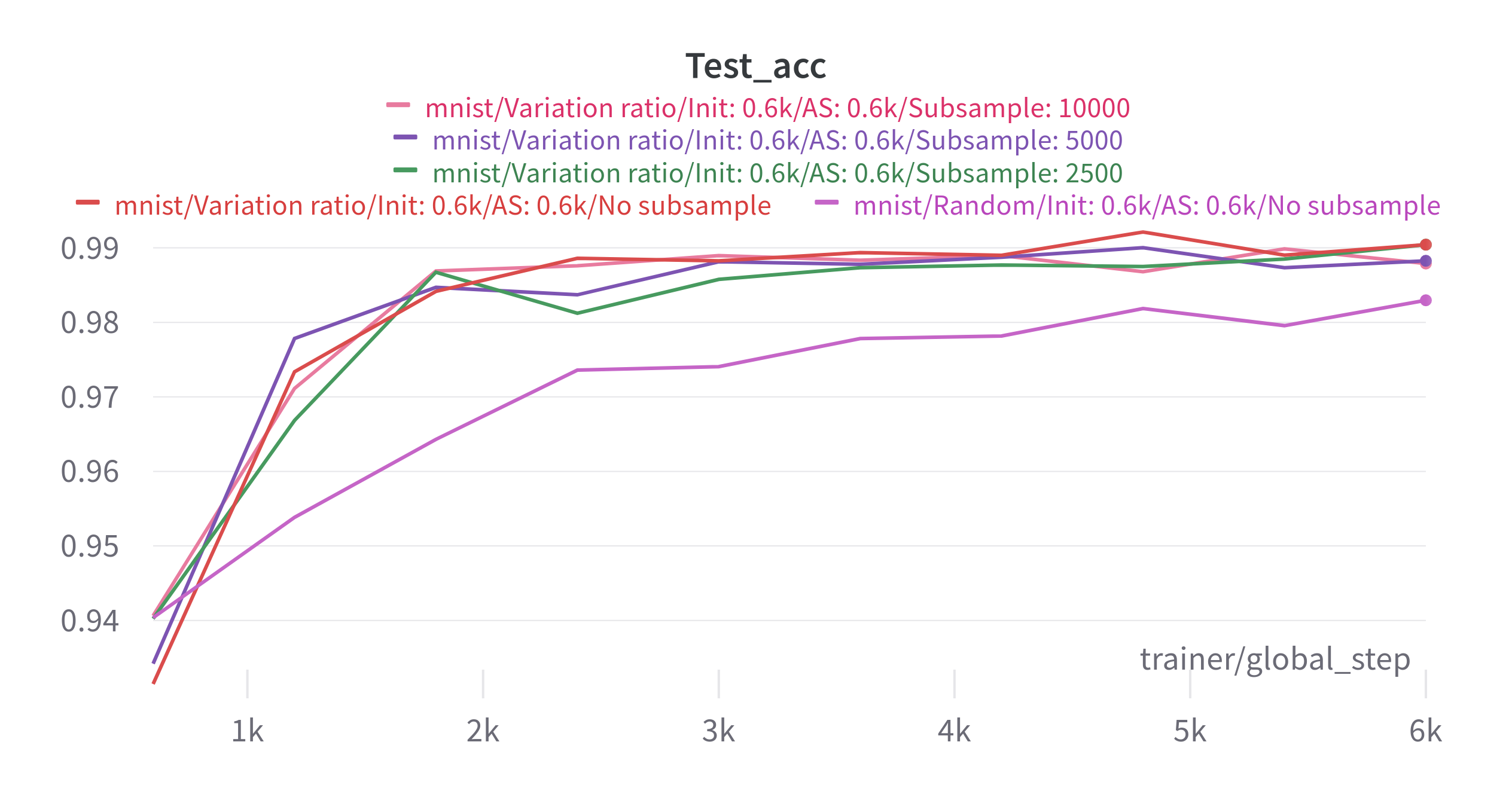}
  \caption{AF: varR, IP: 1\%, TAS: 10\%}
  \label{fig:mnist_var_1_10}
\end{subfigure}%
\begin{subfigure}{.5\textwidth}
  \centering
  \includegraphics[width=1.05\linewidth]{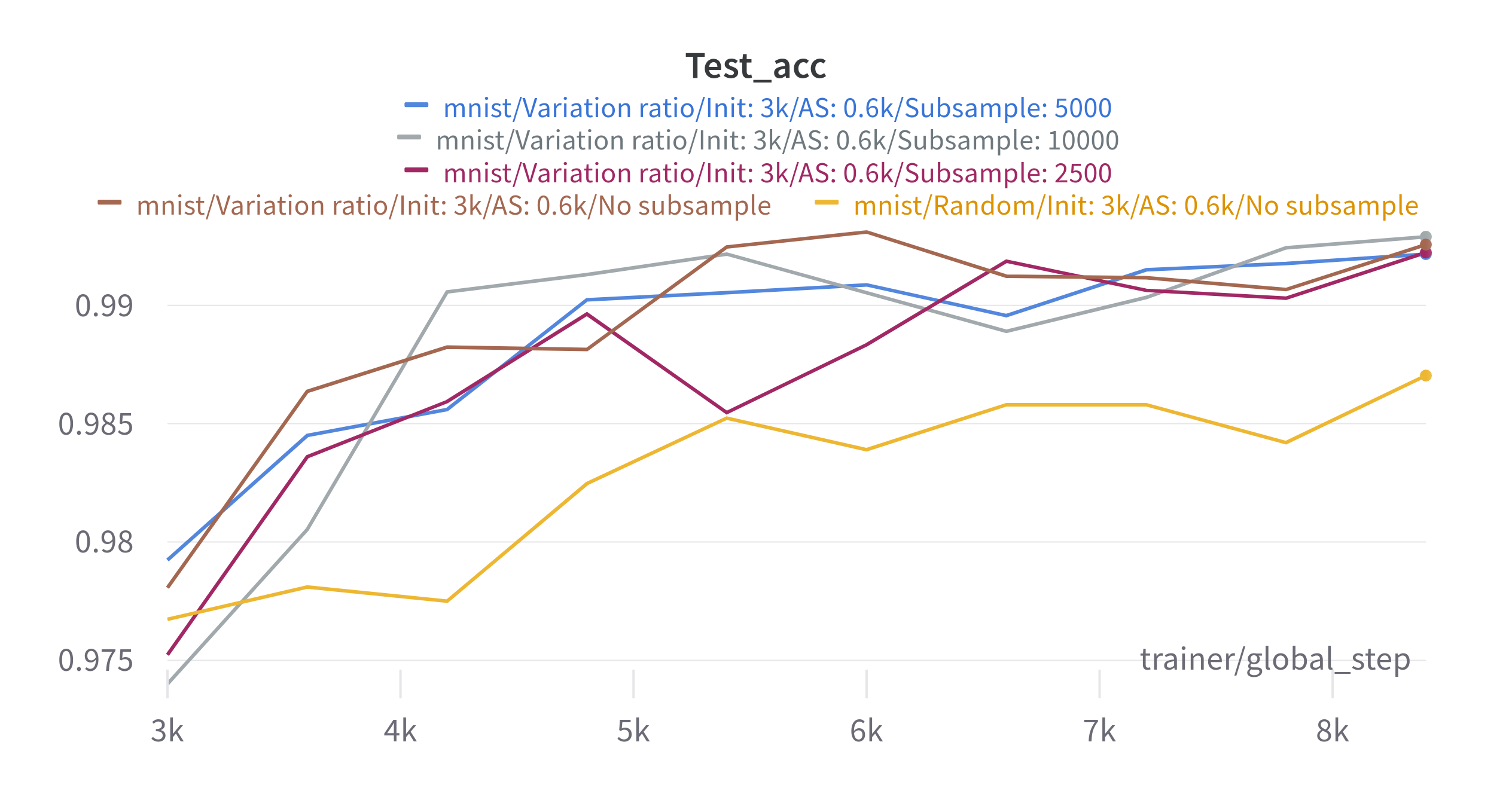}
  \caption{AF: varR, IP: 5\%, TAS: 10\%}
  \label{fig:mnist_var_5_10}
\end{subfigure}
\caption{Test results (classification accuracy) on MNIST dataset with various acquisition functions (AF), initial pool size (IS), and total acquisition size (TAS).}
\label{fig:mnist_exps}
\end{figure}

To assess the effectiveness of our proposed compute-efficient active learning approach, we conducted a series of experiments using the MNIST (\cite{lecun1998gradient}) and CIFAR-10 (\cite{cifar10}) datasets. All experiments shared a common randomly initialized labeled dataset. The class-wise manner balance was also ensured during the initial pool generation. The models employed were consistent across experiments. We utilized acquisition functions based on MC Dropout for entropy and variation ratios calculation, and sampling probabilities were determined by applying the softmax function to the acquisition function values. The number of retraining and sampling iterations $T$ was set to 10. Each model within a retraining iteration underwent training three times with the same dataset using different random seeds, and the results were averaged. We conducted several experiments with various hyperparameter values to investigate the effect of the initial pool size, the total acquisition size, the acquisition function, and the candidate pool size.

For the MNIST experiments, we utilized a simple and compact convolutional neural network architecture with two Conv-MaxPool-Dropout-ReLU layers and a Linear layer. No data augmentations were applied. For further details regarding parameters, please refer to our code. We set the initial pool size to 1\% and 5\% of the training set (60k images), while the total acquisition size is 10\% of the training data. Since the retraining iteration number was 10 during our experiments, we sampled 600 data points at each iteration and added them to the initial pool with the corresponding labels. The $\alpha$ subsampling ratio was set to 2500, 5000, and 10000. We chose classification accuracy as the evaluation metric and computed this value on the test set at the end of each iteration. We used random sampling and entropy-based selection from the whole training set (i.e., no subsampling w.r.t to the candidate pool) as the baseline models. The outcome of the experiments is presented in Figure \ref{fig:mnist_exps}. Our method outperformed the random baseline in every experiment. Furthermore, some of our compute-efficient subsampling methods even outperformed both the entropy and varR-based methods sampled from the whole training set. When the baselines using entropy or varR-based selection performed better than the proposed method, the performance differences were negligible despite the substantial difference concerning compute requirements.

\begin{figure}[t]
\begin{subfigure}{.5\textwidth}
  \centering
  \includegraphics[width=1.05\linewidth]{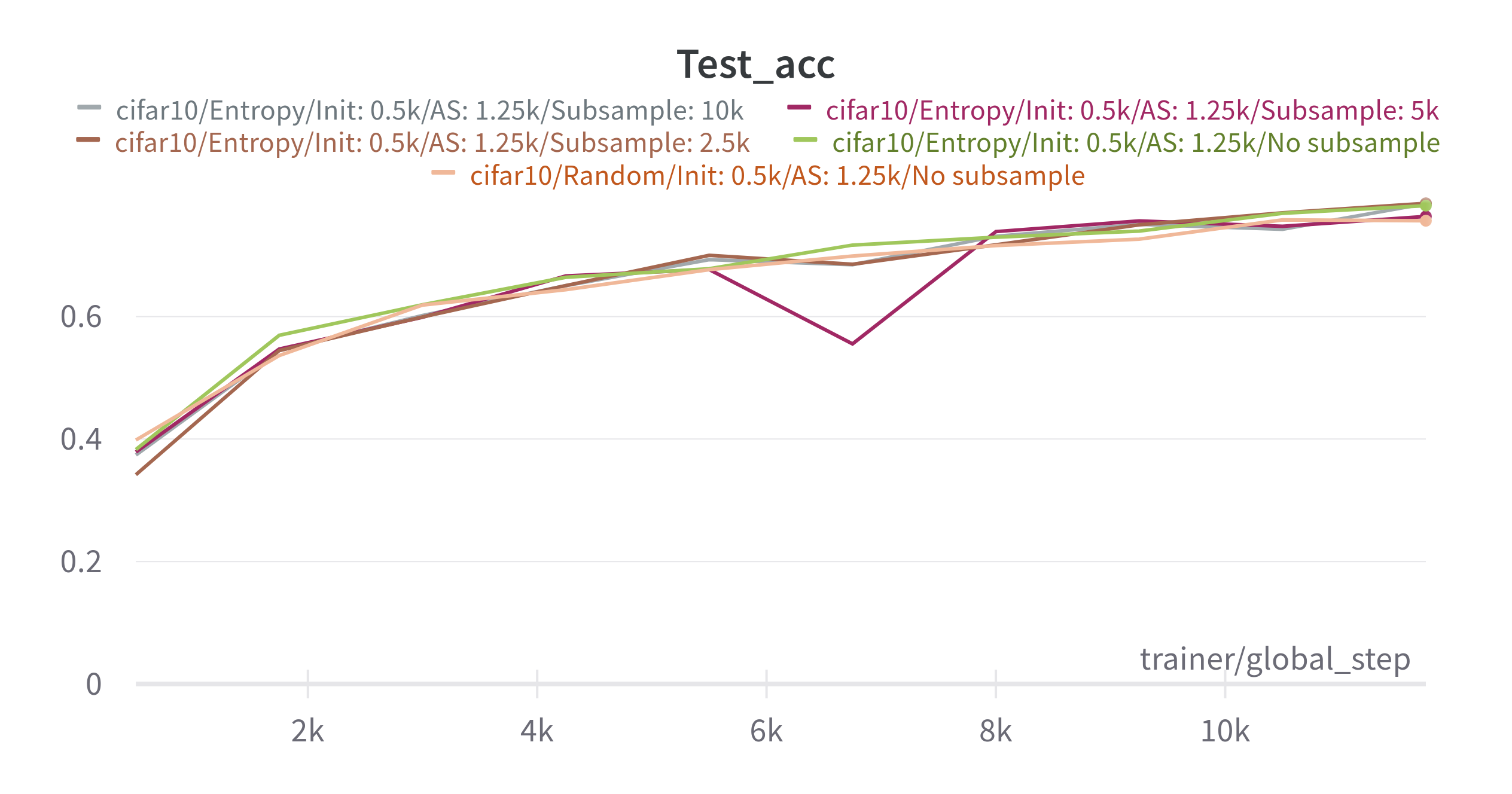}
  \caption{AF: entropy, IP: 1\%, TAS: 25\%}
  \label{fig:cifar_1_25}
\end{subfigure}%
\begin{subfigure}{.5\textwidth}
  \centering
  \includegraphics[width=1.05\linewidth]{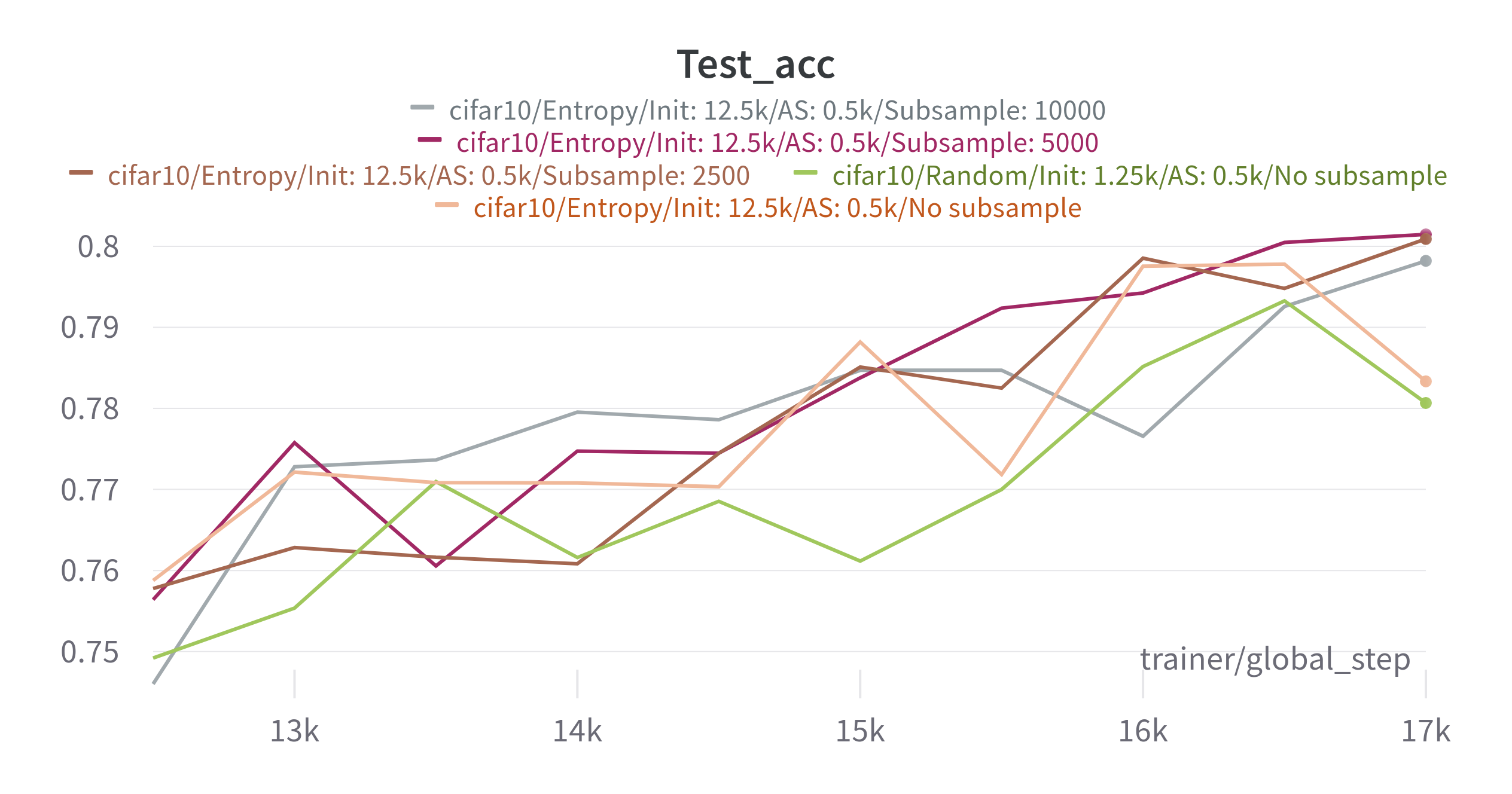}
  \caption{AF: entropy, IP: 25\%, TAS: 10\%}
  \label{fig:cifar_25_10}
\end{subfigure}
\begin{subfigure}{.5\textwidth}
  \centering
  \includegraphics[width=1.05\linewidth]{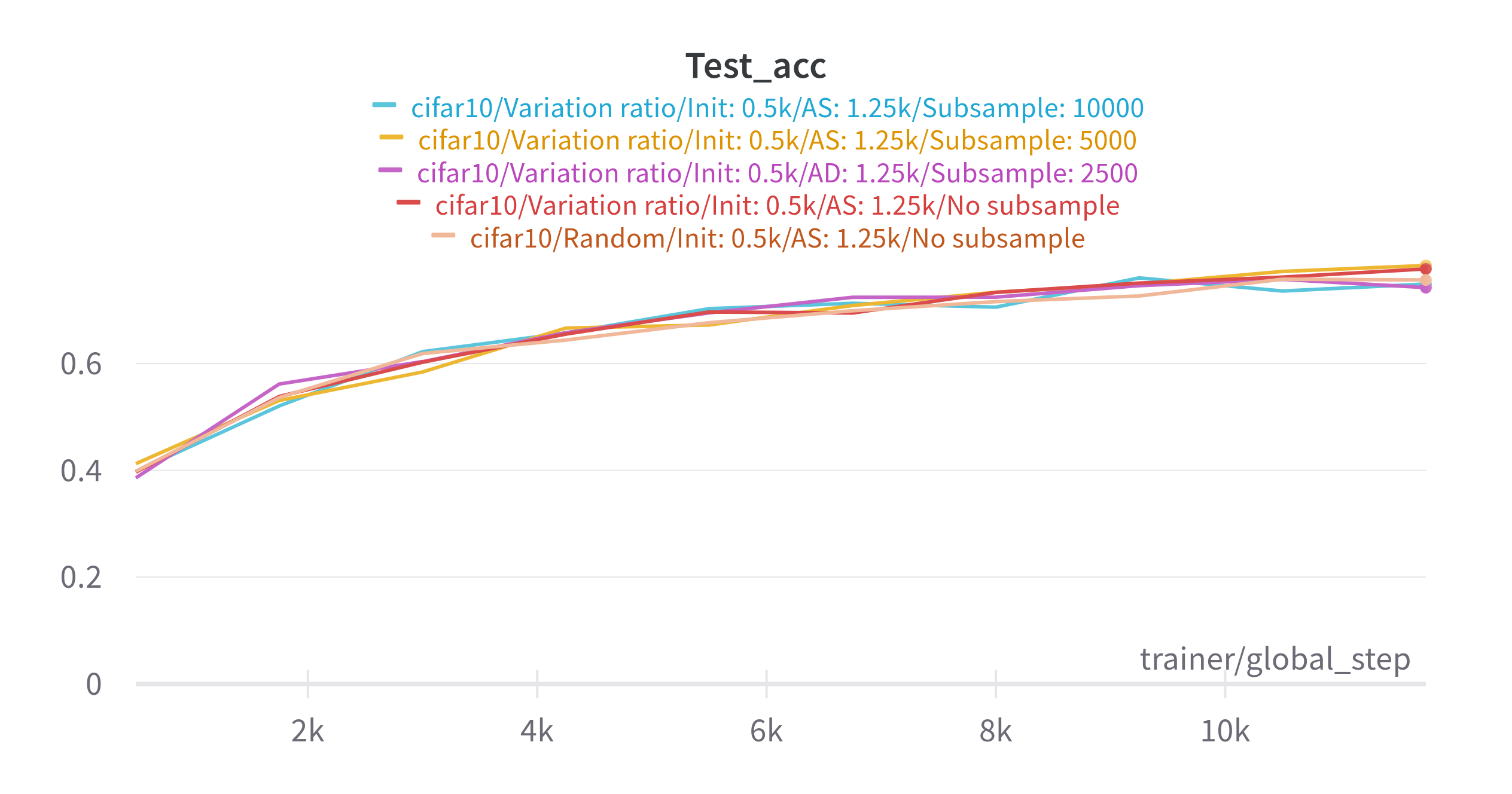}
  \caption{AF: varR, IP: 1\%, TAS: 25\%}
  \label{fig:cifar_var_1_25}
\end{subfigure}%
\begin{subfigure}{.5\textwidth}
  \centering
  \includegraphics[width=1.05\linewidth]{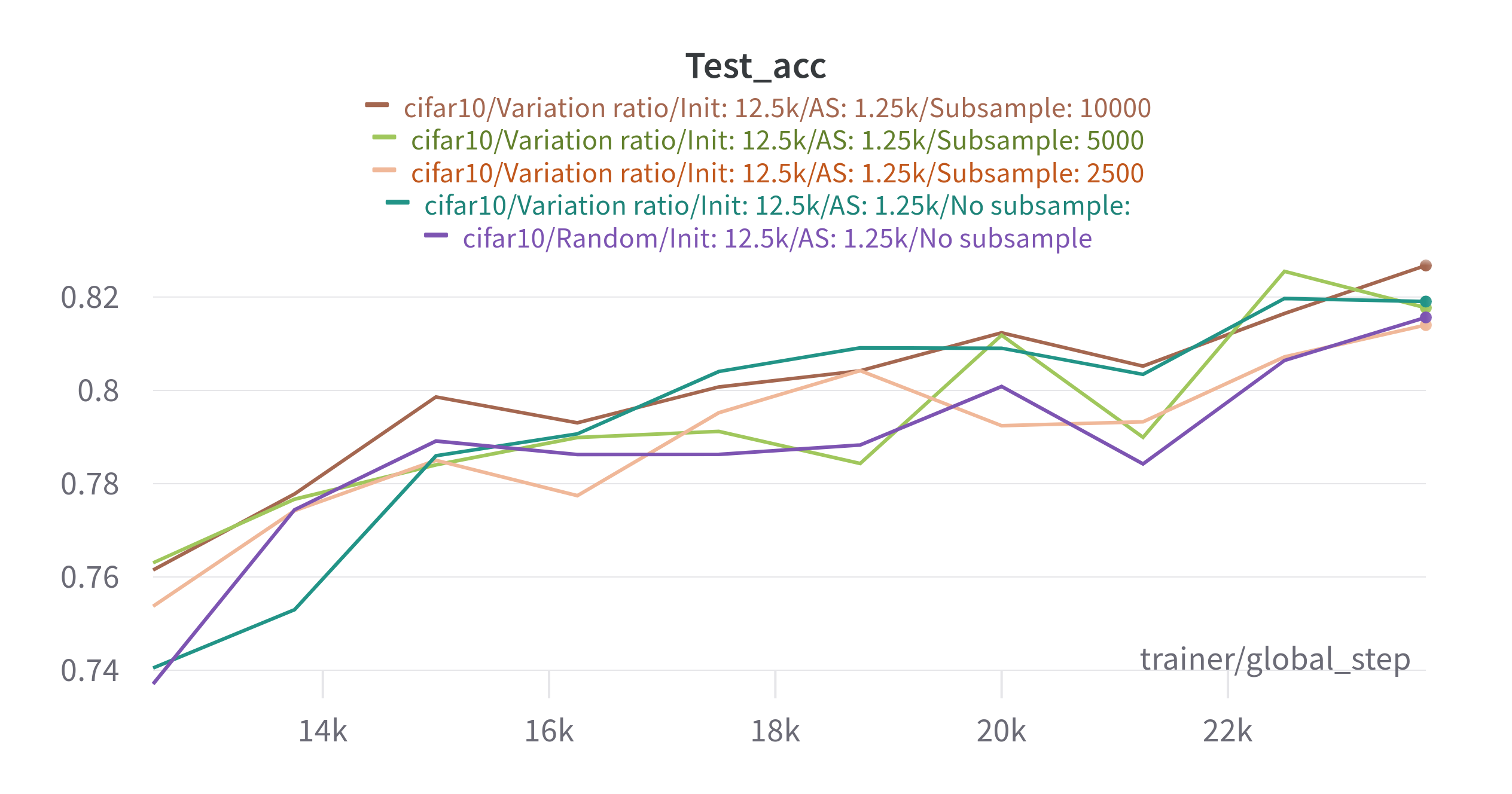}
  \caption{AF: varR, IP: 25\%, TAS: 25\%}
  \label{fig:cifar_var_25_25}
\end{subfigure}
\caption{Test results (classification accuracy) on CIFAR-10 dataset with various acquisition functions (AF), initial pool size (IS), and total acquisition size (TAS).}
\label{fig:cifar_exps}
\end{figure}

Since handwritten digit classification is an easy problem and the simple CNN model was able to achieve good performance (99\% classification accuracy) using only 6\% of the whole dataset for training and 20\% for the candidate pool size, we conducted experiments using the more complex CIFAR-10 dataset. We selected VGG-11 (\cite{simonyan2014very}) with batch normalization as the architecture. In contrast to the MNIST experiments, we did apply data augmentations, namely random brightness contrast and horizontal flip (please refer to our code for finding the hyperparameter values). The active learning setup was similar to the MNIST case. The initial pool (1\% and 25\%) and the total acquisition size (10\% and 25\%, 500 and 1250 data points per iteration, respectively) used for the experiments have been changed. Other settings were not modified. Like in the MNIST experiments, both baselines were outperformed again by our proposed method. While the performance difference was more visible in the small data regime (initial pool size: 1\%) in the case of the MNIST dataset, our method performs better on CIFAR-10 when the initial set is larger (initial pool size: 25\%). This fact might be explained by the complexity of the dataset in comparison to MNIST.  In terms of compute efficiency (i.e., training time), the proposed method can save up to 25\% (502 min vs. 672 min trained on an NVIDIA GeForce GTX TITAN X) using 26\% of the entire training set (initial pool size: 1\%, total acquisition size: 25\%) while still outperforming the baseline (variation ratios-based sampling from the whole training set, 78.36\% vs. 77.71\% classification accuracy). Since the ratio of candidate pool and entire training set size is significantly smaller in the case of large-scale real-world datasets than in our experiments using the CIFAR-10 dataset, we assume the runtime-decreasing phenomenon is even more visible. However, additional experiments have to be conducted to provide quantitative numbers and validate this hypothesis.

%\begin{figure}
%    \centering
%    \includegraphics[width=1.0\linewidth]{mnist.png}
%    \caption{The outcomes of our approach on the MNIST dataset were examined across different subsampling rates. Our findings demonstrate that subsampling not only decreases computational requirements but can also yield %improvements in predictive accuracy, especially during the initial phases of active learning.}
%    \label{fig:mnist}
%\end{figure}

As the experimental results show, despite not utilizing all available samples within the active learning process, the application of our proposed framework yielded notable improvements. This was not only marked by a reduction in computational demands but also by an enhancement in predictive performance. One of the underlying hypotheses for this phenomenon lies in the fact that acquisition functions often tend to select less diverse samples, as they prioritize instances about which the model exhibits uncertainty. In essence, our method can be viewed as an interpolation between random sampling and acquisition function-based sampling where the temperature parameter of the softmax function controls the interpolation. This aspect holds particular significance, especially in the initial stages of the active learning cycle, where random sampling serves as a robust baseline. Additionally, our approach addresses the cold start problem encountered in active learning, where acquisition function-based methods struggle when confronted with small initial datasets.

To demonstrate the model and task agnosticism of the proposed method, we are conducting experiments with multimodal 3D object detection on aiMotive Multimodal Dataset (\cite{matuszka2023aimotive}) using a BEVFusion-like model (\cite{liu2023bevfusion}). The MC Dropout method can also be used as the acquisition function for 3D object detection where the entropy of the localization was utilized as the acquisition function value. The preliminary results showed a similar phenomenon as Figure \ref{fig:cifar_exps} depicts, indicating the versatility of the proposed solution and its usability for complex real-world tasks using a larger model than described in the first two groups of experiments. 

\section{Conclusion}
In this study, we have introduced an innovative framework designed to alleviate the computational demands associated with acquisition function-based active learning techniques. Our approach leverages a candidate dataset for informed sampling, grounded in the premise that historical acquisition function values serve as reliable predictors of future values. We have empirically validated the efficacy of our method on established benchmark datasets, demonstrating its practical utility. It's worth noting that while these benchmark datasets provide well-defined problem settings, our internal experiments further show the effectiveness of our approach. Despite the limitations inherent to benchmark datasets, our results affirm the promise and adaptability of our method in addressing the broader challenges of active learning.

% Acknowledgements should go at the end, before appendices and references

%\acks{We would like to acknowledge support for this project
%from the National Science Foundation (NSF grant IIS-9988642)
%and the Multidisciplinary Research Program of the Department
%of Defense (MURI N00014-00-1-0637). }


\bibliography{sample}

\begin{thebibliography}{14}
\providecommand{\natexlab}[1]{#1}
\providecommand{\url}[1]{\texttt{#1}}
\expandafter\ifx\csname urlstyle\endcsname\relax
  \providecommand{\doi}[1]{doi: #1}\else
  \providecommand{\doi}{doi: \begingroup \urlstyle{rm}\Url}\fi

\bibitem[Beluch et~al.(2018)Beluch, Genewein, Nürnberger, and Köhler]{al_ensemble}
William~H. Beluch, Tim Genewein, Andreas Nürnberger, and Jan~M. Köhler.
\newblock The power of ensembles for active learning in image classification.
\newblock In \emph{Proceedings of the IEEE Conference on Computer Vision and Pattern Recognition (CVPR)}, June 2018.

\bibitem[Gal and Ghahramani(2015)]{mc_dropout}
Yarin Gal and Zoubin Ghahramani.
\newblock Dropout as a bayesian approximation: Representing model uncertainty in deep learning, 2015.
\newblock URL \url{https://arxiv.org/abs/1506.02142}.

\bibitem[Gal et~al.(2017)Gal, Islam, and Ghahramani]{al_bald}
Yarin Gal, Riashat Islam, and Zoubin Ghahramani.
\newblock Deep {B}ayesian active learning with image data.
\newblock In Doina Precup and Yee~Whye Teh, editors, \emph{Proceedings of the 34th International Conference on Machine Learning}, volume~70 of \emph{Proceedings of Machine Learning Research}, pages 1183--1192. PMLR, 06--11 Aug 2017.
\newblock URL \url{https://proceedings.mlr.press/v70/gal17a.html}.

\bibitem[Krizhevsky et~al.(2009)Krizhevsky, Hinton, et~al.]{cifar10}
Alex Krizhevsky, Geoffrey Hinton, et~al.
\newblock Learning multiple layers of features from tiny images.
\newblock 2009.

\bibitem[LeCun et~al.(1998)LeCun, Bottou, Bengio, and Haffner]{lecun1998gradient}
Yann LeCun, L{\'e}on Bottou, Yoshua Bengio, and Patrick Haffner.
\newblock Gradient-based learning applied to document recognition.
\newblock \emph{Proceedings of the IEEE}, 86\penalty0 (11):\penalty0 2278--2324, 1998.

\bibitem[Liu et~al.(2023)Liu, Tang, Amini, Yang, Mao, Rus, and Han]{liu2023bevfusion}
Zhijian Liu, Haotian Tang, Alexander Amini, Xinyu Yang, Huizi Mao, Daniela~L Rus, and Song Han.
\newblock Bevfusion: Multi-task multi-sensor fusion with unified bird's-eye view representation.
\newblock In \emph{2023 IEEE International Conference on Robotics and Automation (ICRA)}, pages 2774--2781. IEEE, 2023.

\bibitem[Matuszka et~al.(2023)Matuszka, Barton, Butykai, Hajas, Kiss, Kov{\'a}cs, Kuns{\'a}gi-M{\'a}t{\'e}, Lengyel, N{\'e}meth, Pet{\H{o}}, et~al.]{matuszka2023aimotive}
Tamas Matuszka, Ivan Barton, {\'A}d{\'a}m Butykai, P{\'e}ter Hajas, D{\'a}vid Kiss, Domonkos Kov{\'a}cs, S{\'a}ndor Kuns{\'a}gi-M{\'a}t{\'e}, P{\'e}ter Lengyel, G{\'a}bor N{\'e}meth, Levente Pet{\H{o}}, et~al.
\newblock aimotive dataset: A multimodal dataset for robust autonomous driving with long-range perception.
\newblock In \emph{International Conference on Learning Representations 2023 Workshop on Scene Representations for Autonomous Driving}, 2023.

\bibitem[Park et~al.(2023)Park, Choi, Kim, Han, and Moon]{park2023active}
Younghyun Park, Wonjeong Choi, Soyeong Kim, Dong-Jun Han, and Jaekyun Moon.
\newblock Active learning for object detection with evidential deep learning and hierarchical uncertainty aggregation.
\newblock In \emph{The Eleventh International Conference on Learning Representations}, 2023.
\newblock URL \url{https://openreview.net/forum?id=MnEjsw-vj-X}.

\bibitem[Ren et~al.(2021)Ren, Xiao, Chang, Huang, Li, Gupta, Chen, and Wang]{al_survey}
Pengzhen Ren, Yun Xiao, Xiaojun Chang, Po-Yao Huang, Zhihui Li, Brij~B. Gupta, Xiaojiang Chen, and Xin Wang.
\newblock A survey of deep active learning.
\newblock \emph{ACM Comput. Surv.}, 54\penalty0 (9), oct 2021.
\newblock ISSN 0360-0300.
\newblock \doi{10.1145/3472291}.
\newblock URL \url{https://doi.org/10.1145/3472291}.

\bibitem[Senay et~al.(2020)Senay, Idrissi, and Haziza]{al_vat}
Gregory Senay, Badr~Youbi Idrissi, and Marine Haziza.
\newblock Viraal: Virtual adversarial active learning.
\newblock \emph{CoRR}, 2020.

\bibitem[Simonyan and Zisserman(2014)]{simonyan2014very}
Karen Simonyan and Andrew Zisserman.
\newblock Very deep convolutional networks for large-scale image recognition.
\newblock \emph{arXiv preprint arXiv:1409.1556}, 2014.

\bibitem[Sinha et~al.(2019)Sinha, Ebrahimi, and Darrell]{al_vaal}
Samarth Sinha, Sayna Ebrahimi, and Trevor Darrell.
\newblock Variational adversarial active learning, 2019.
\newblock URL \url{https://arxiv.org/abs/1904.00370}.

\bibitem[Yi et~al.(2022)Yi, Seo, Park, and Choi]{pt4al}
John Seon~Keun Yi, Minseok Seo, Jongchan Park, and Dong-Geol Choi.
\newblock Pt4al: Using self-supervised pretext tasks for active learning.
\newblock In \emph{European Conference on Computer Vision}, pages 596--612. Springer, 2022.

\bibitem[Yoo and Kweon(2019)]{learning_loss}
Donggeun Yoo and In~So Kweon.
\newblock Learning loss for active learning, 2019.
\newblock URL \url{https://arxiv.org/abs/1905.03677}.

\end{thebibliography}

\newpage

\begin{center}
\textbf{\LARGE Appendix - Compute-Efficient Active Learning}
\end{center}

%------------------------------------------------------------------------
\setcounter{section}{0}

\section{Additional experimental results}
\label{sup_mat_exps}
\begin{figure}[t]
\begin{subfigure}{.5\textwidth}
  \centering
  \includegraphics[width=1.0\linewidth]{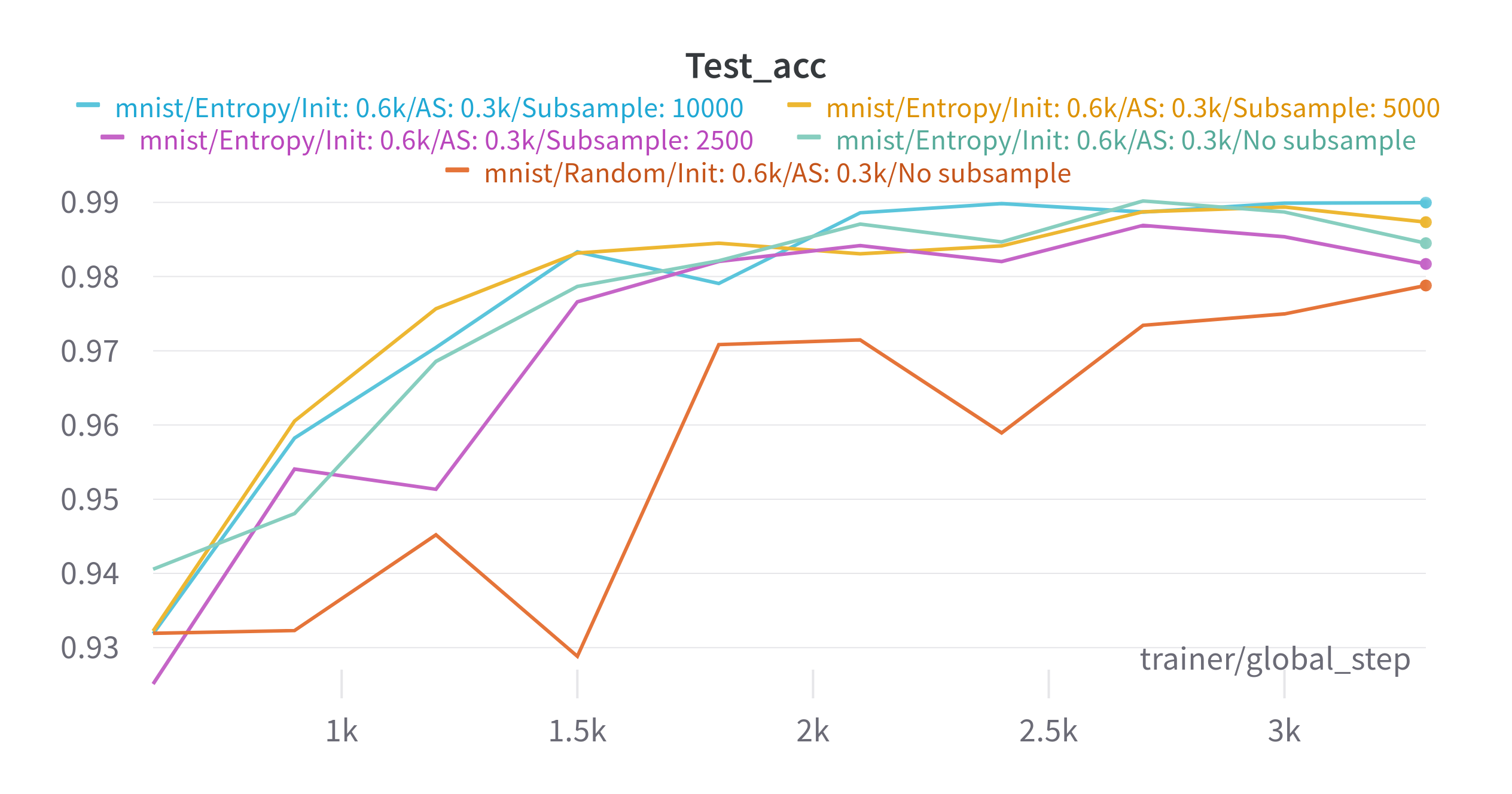}
  \caption{AF: entropy, IP: 1\%, TAS: 5\%}
  \label{fig:mnist_1_5}
\end{subfigure}%
\begin{subfigure}{.5\textwidth}
  \centering
  \includegraphics[width=1.0\linewidth]{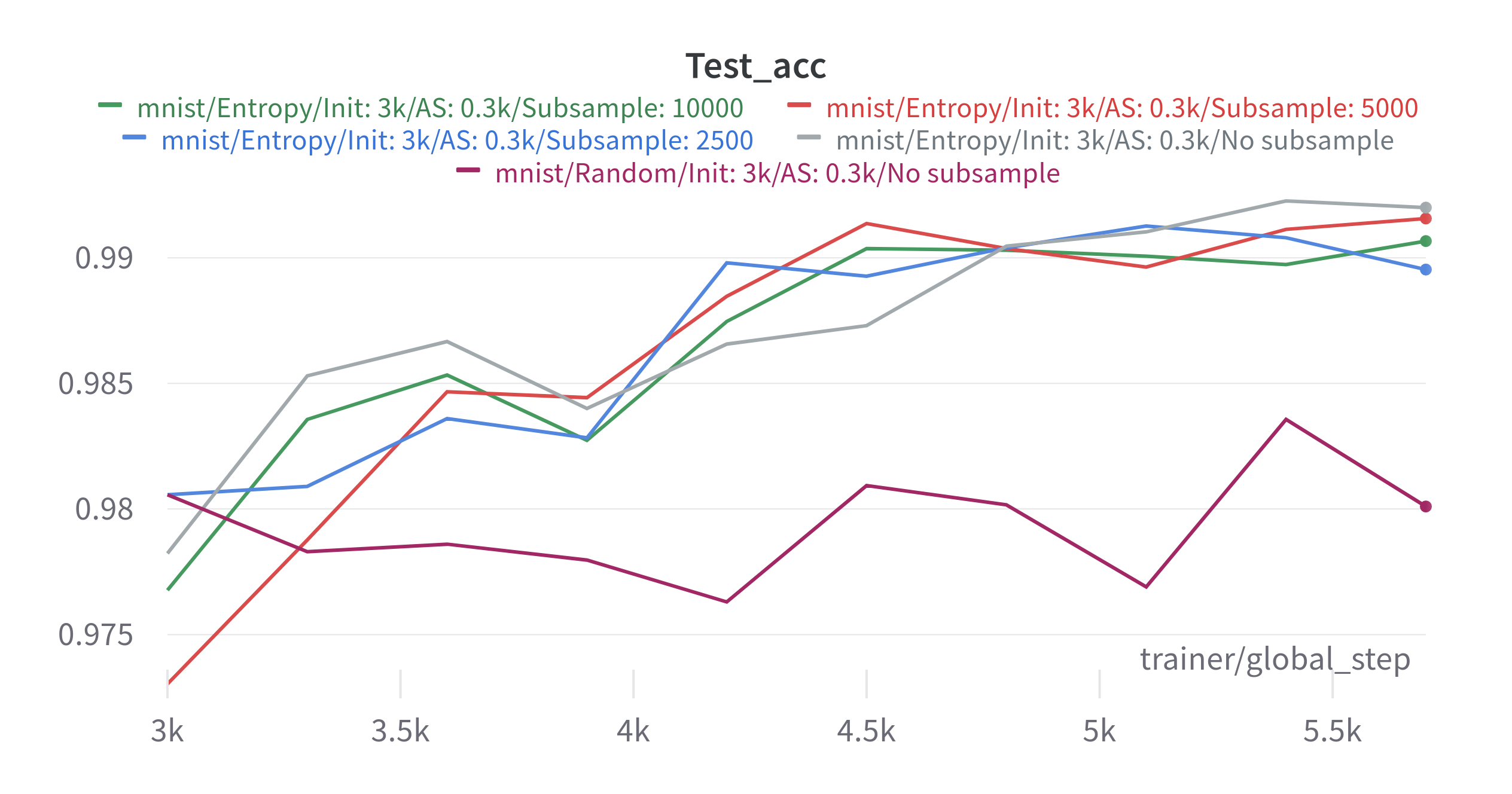}
  
  \caption{AF: entropy, IP: 5\%, TAS: 5\%}
  \label{fig:mnist_5_5}
\end{subfigure}
\begin{subfigure}{.5\textwidth}
  \centering
  \includegraphics[width=1.0\linewidth]{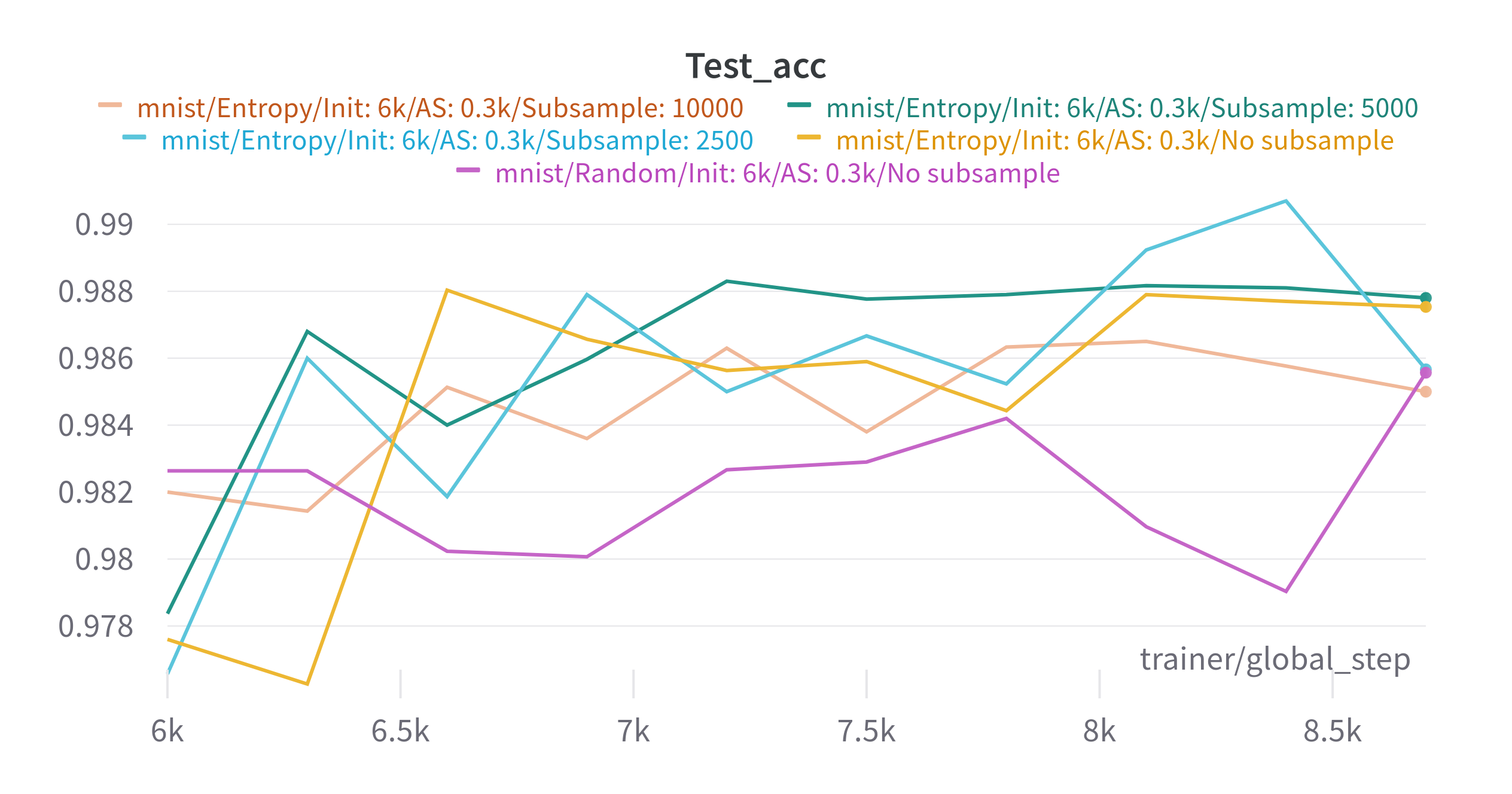}
  \caption{AF: entropy, IP: 10\%, TAS: 5\%}
  \label{fig:mnist_entr_10_5}
\end{subfigure}%
\begin{subfigure}{.5\textwidth}
  \centering
  \includegraphics[width=1.0\linewidth]{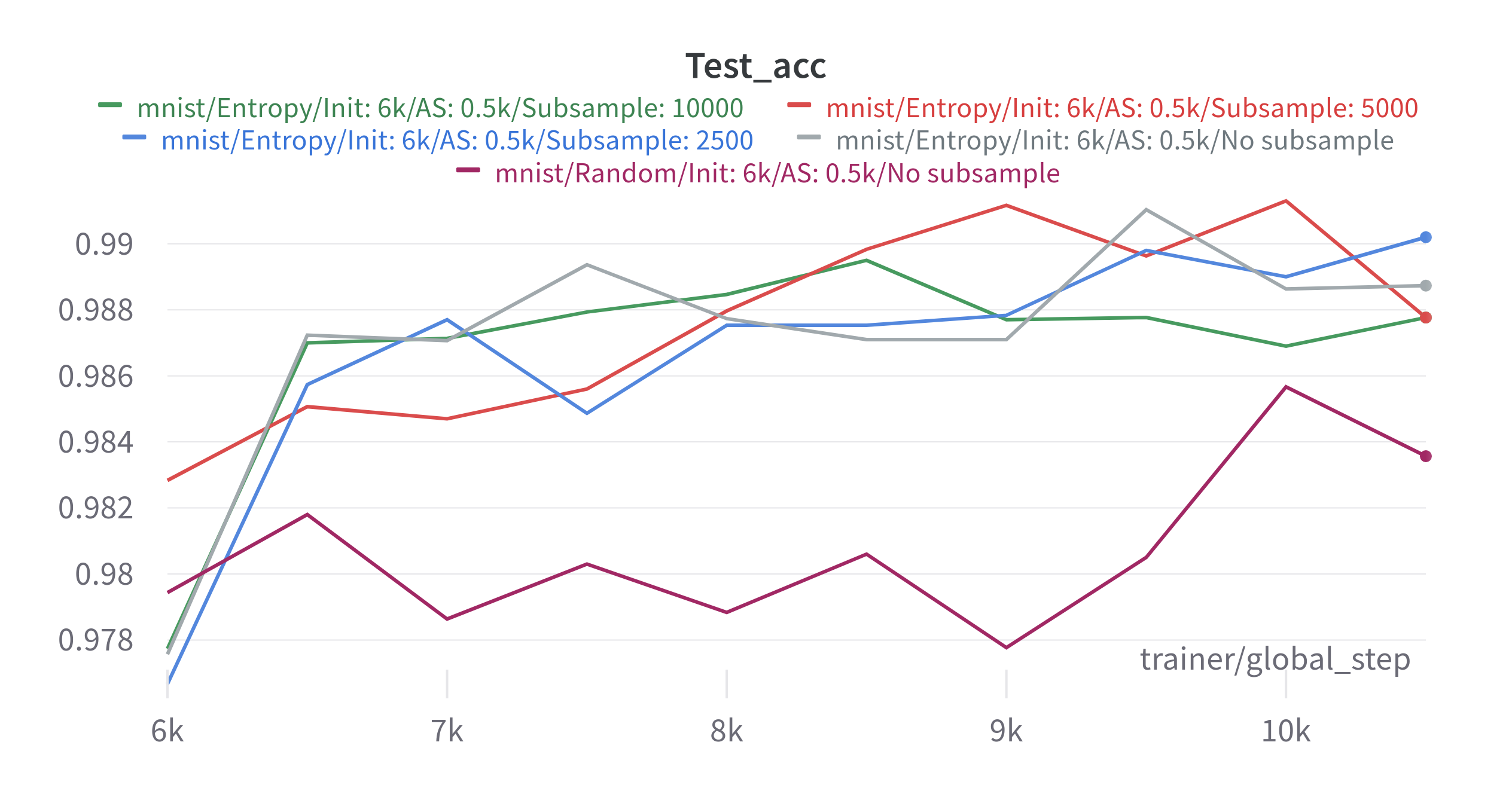}
  \caption{AF: entropy, IP: 10\%, TAS: 8\%}
  \label{fig:mnist_entr_10_8}
\end{subfigure}
\caption{Test results (classification accuracy) on MNIST dataset with entropy acquisition function (AF), various initial pool sizes (IS), and total acquisition sizes (TAS).}
\label{fig:mnist_exps2}
\end{figure}

\begin{figure}[b]
\begin{subfigure}{.5\textwidth}
  \centering
  \includegraphics[width=1.0\linewidth]{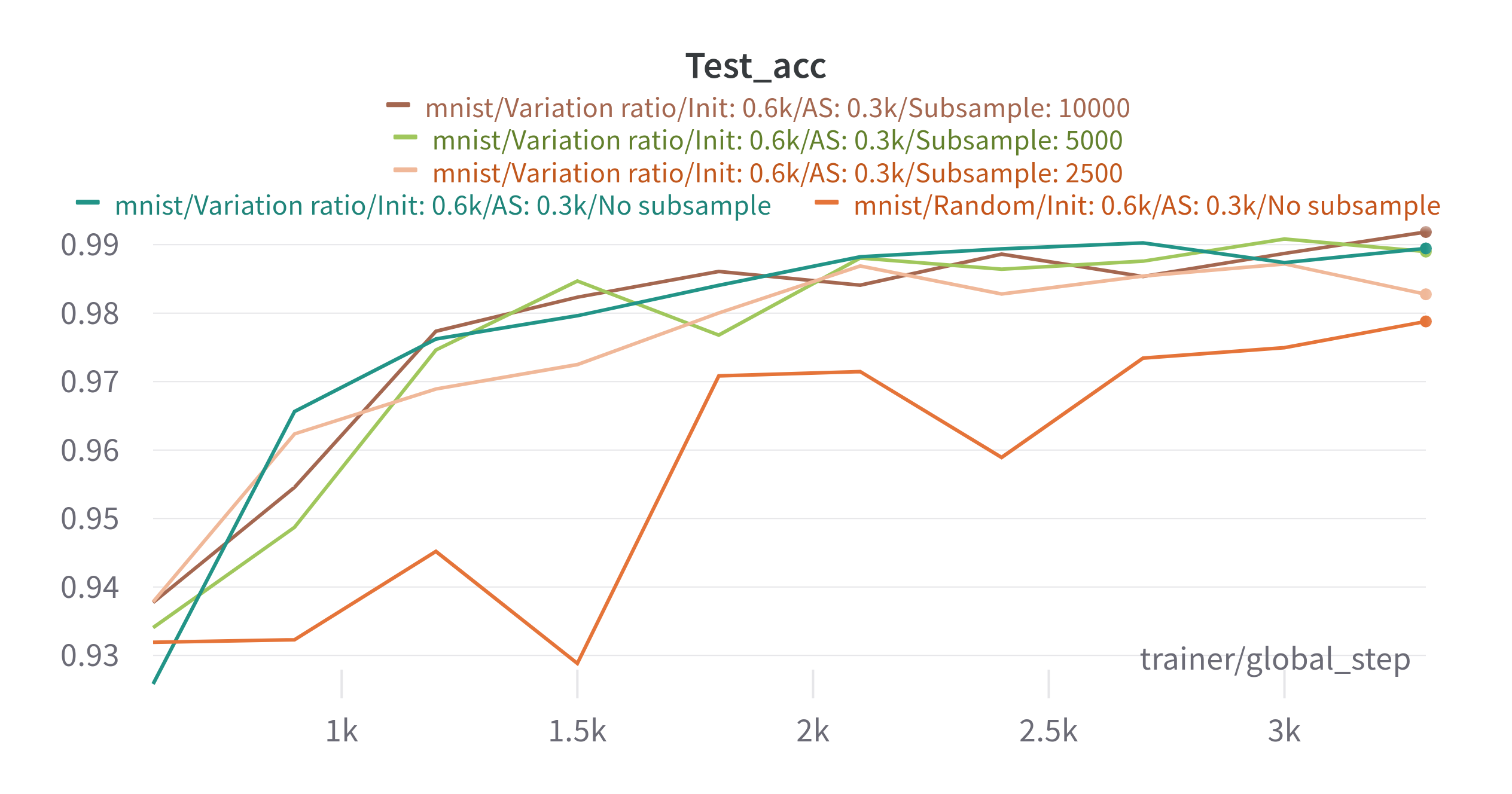}
  \caption{AF: varR, IP: 1\%, TAS: 5\%}
  \label{fig:mnist_var_1_5}
\end{subfigure}%
\begin{subfigure}{.5\textwidth}
  \centering
  \includegraphics[width=1.0\linewidth]{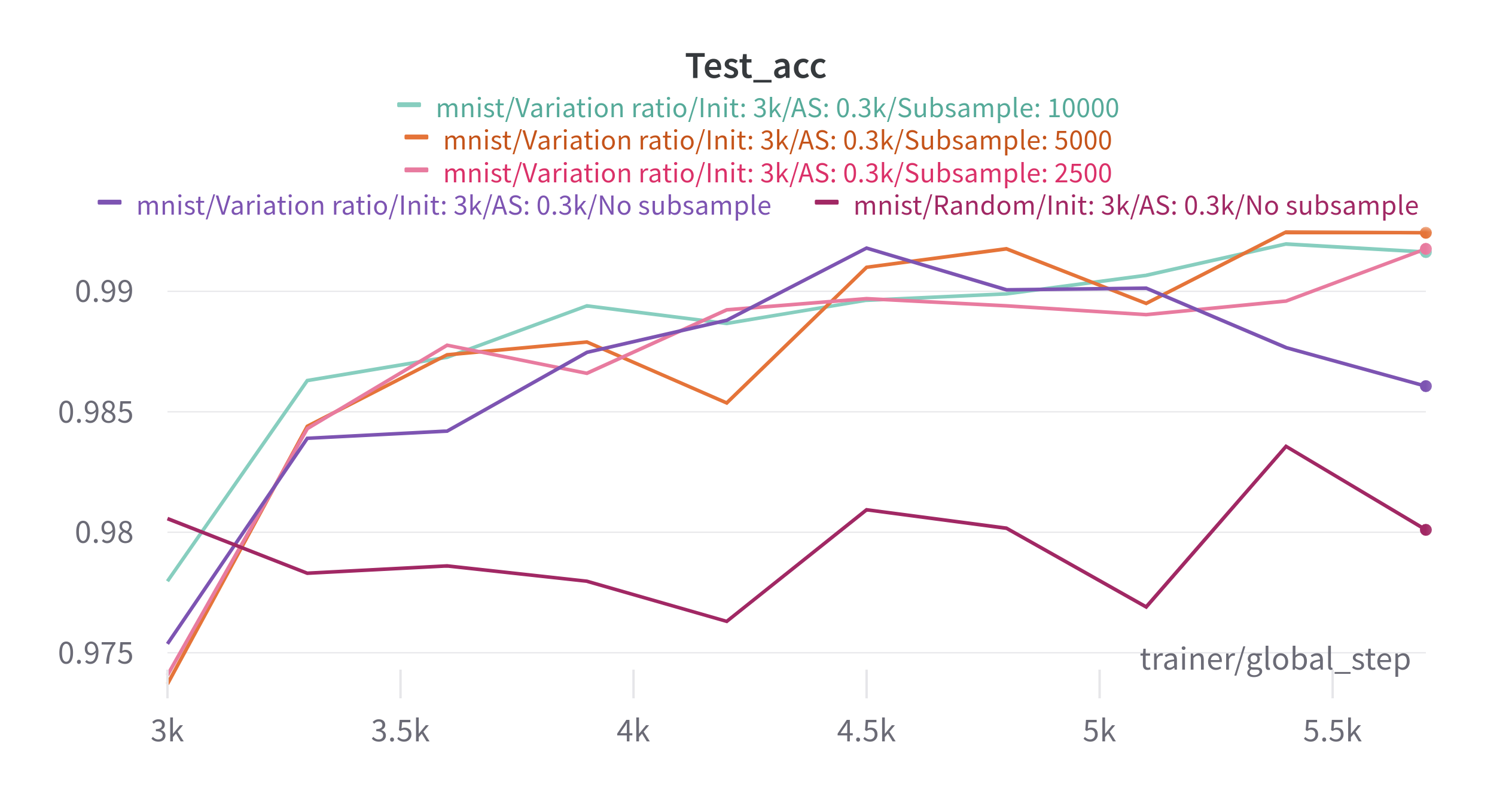}
  
  \caption{AF: varR, IP: 5\%, TAS: 5\%}
  \label{fig:mnist_var_5_5}
\end{subfigure}
\begin{subfigure}{.5\textwidth}
  \centering
  \includegraphics[width=1.0\linewidth]{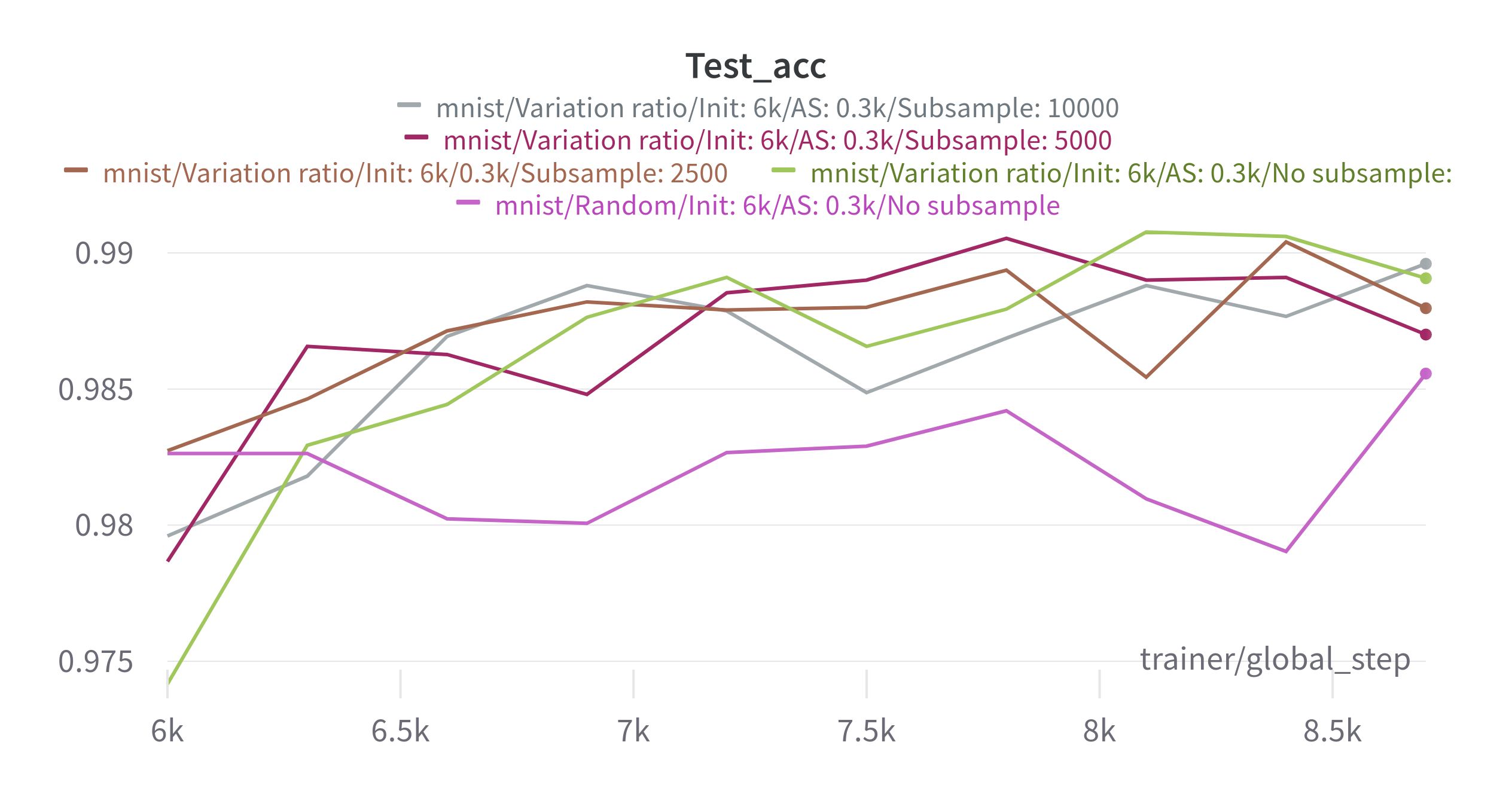}
  \caption{AF: varR, IP: 10\%, TAS: 5\%}
  \label{fig:mnist_var_10_5}
\end{subfigure}%
\begin{subfigure}{.5\textwidth}
  \centering
  \includegraphics[width=1.0\linewidth]{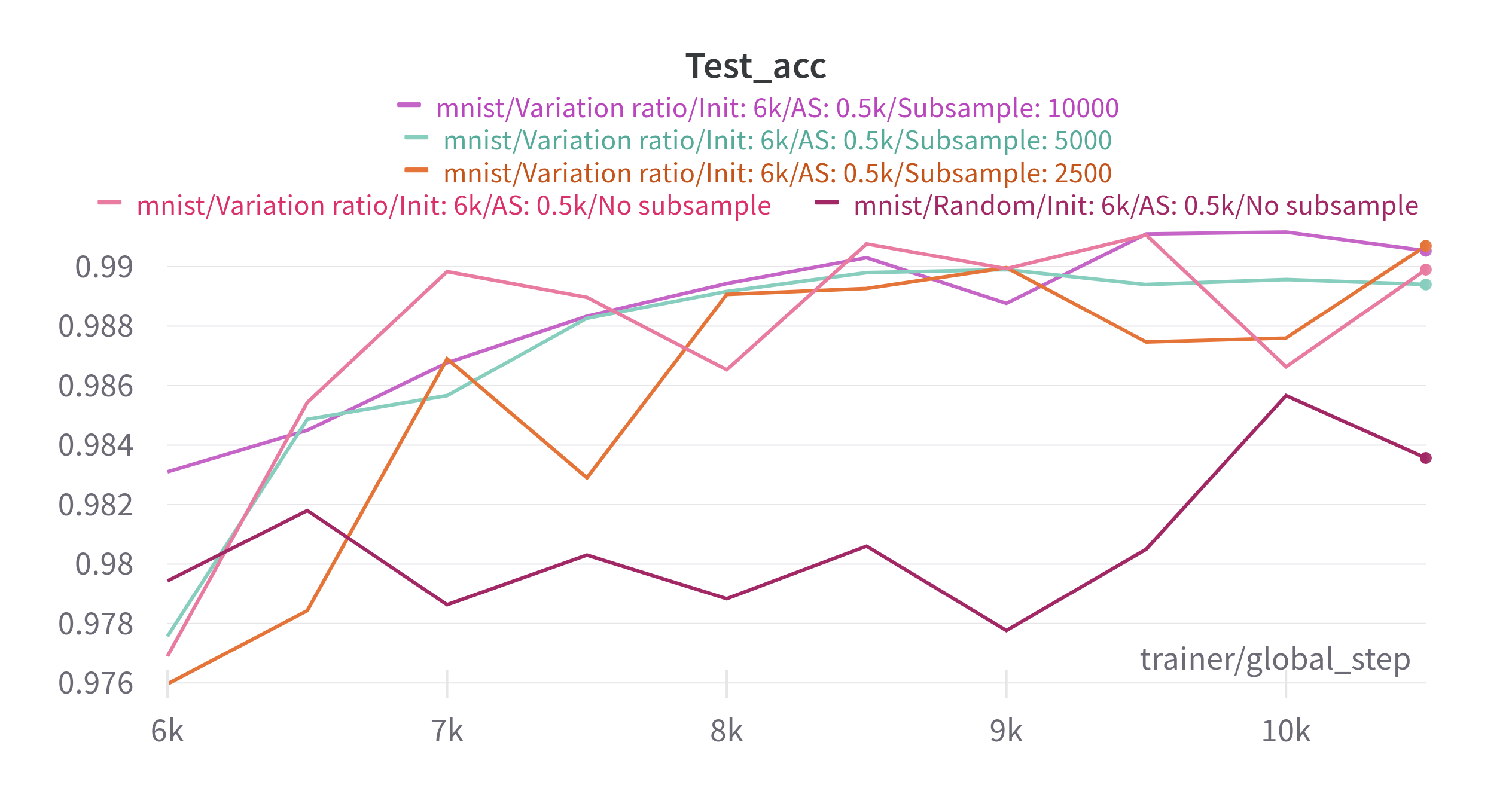}
  \caption{AF: varR, IP: 10\%, TAS: 10\%}
  \label{fig:mnist_var_10_10}
\end{subfigure}
\caption{Test results (classification accuracy) on MNIST dataset with variation ratios acquisition function (AF), various initial pool sizes (IS), and total acquisition sizes (TAS).}
\label{fig:mnist_exps3}
\end{figure}

\begin{figure}[t]
\begin{subfigure}{.5\textwidth}
  \centering
  \includegraphics[width=1.0\linewidth]{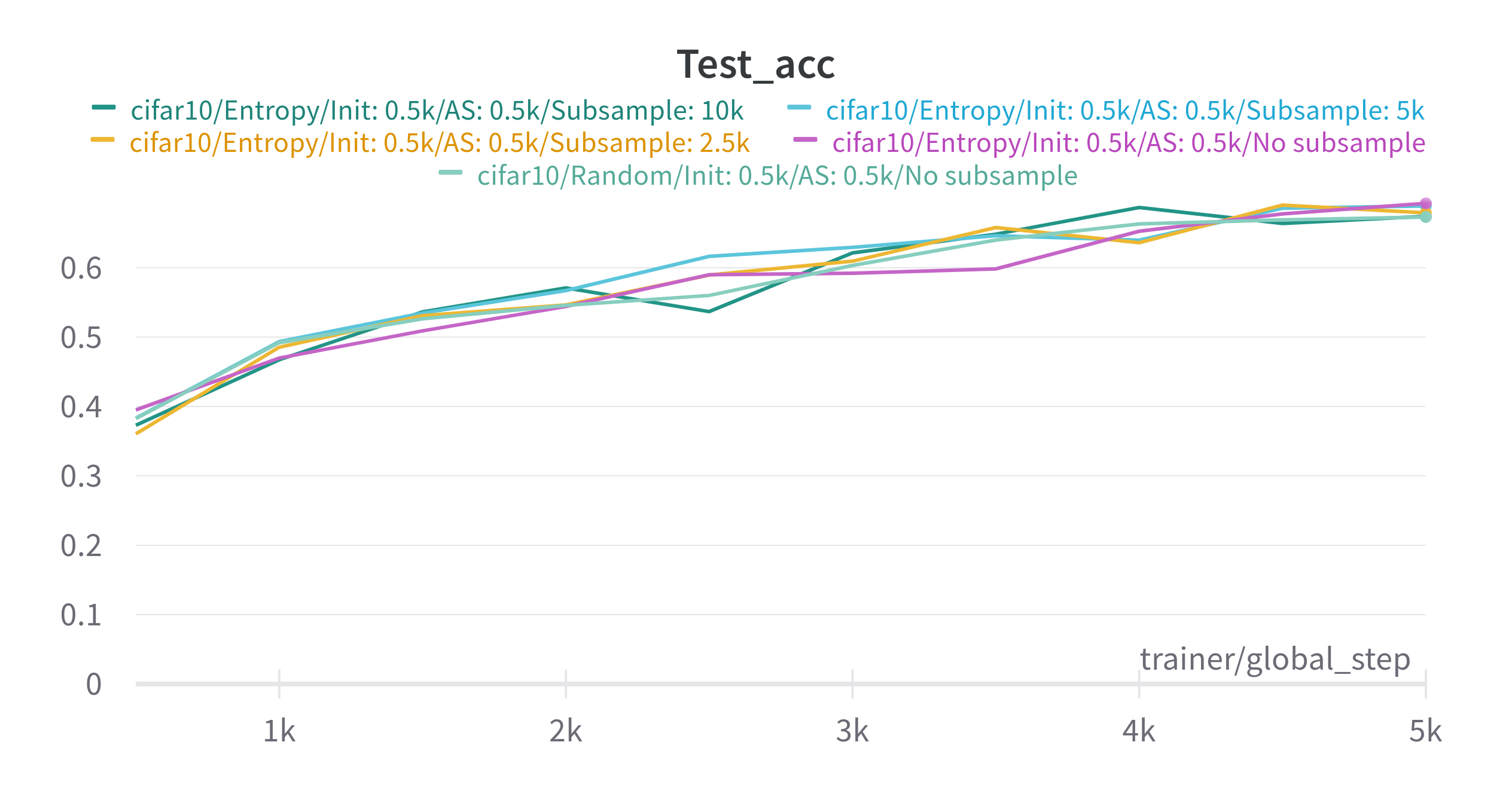}
  \caption{AF: entropy, IP: 1\%, TAS: 10\%}
  \label{fig:cifar_1_10}
\end{subfigure}%
\begin{subfigure}{.5\textwidth}
  \centering
  \includegraphics[width=1.0\linewidth]{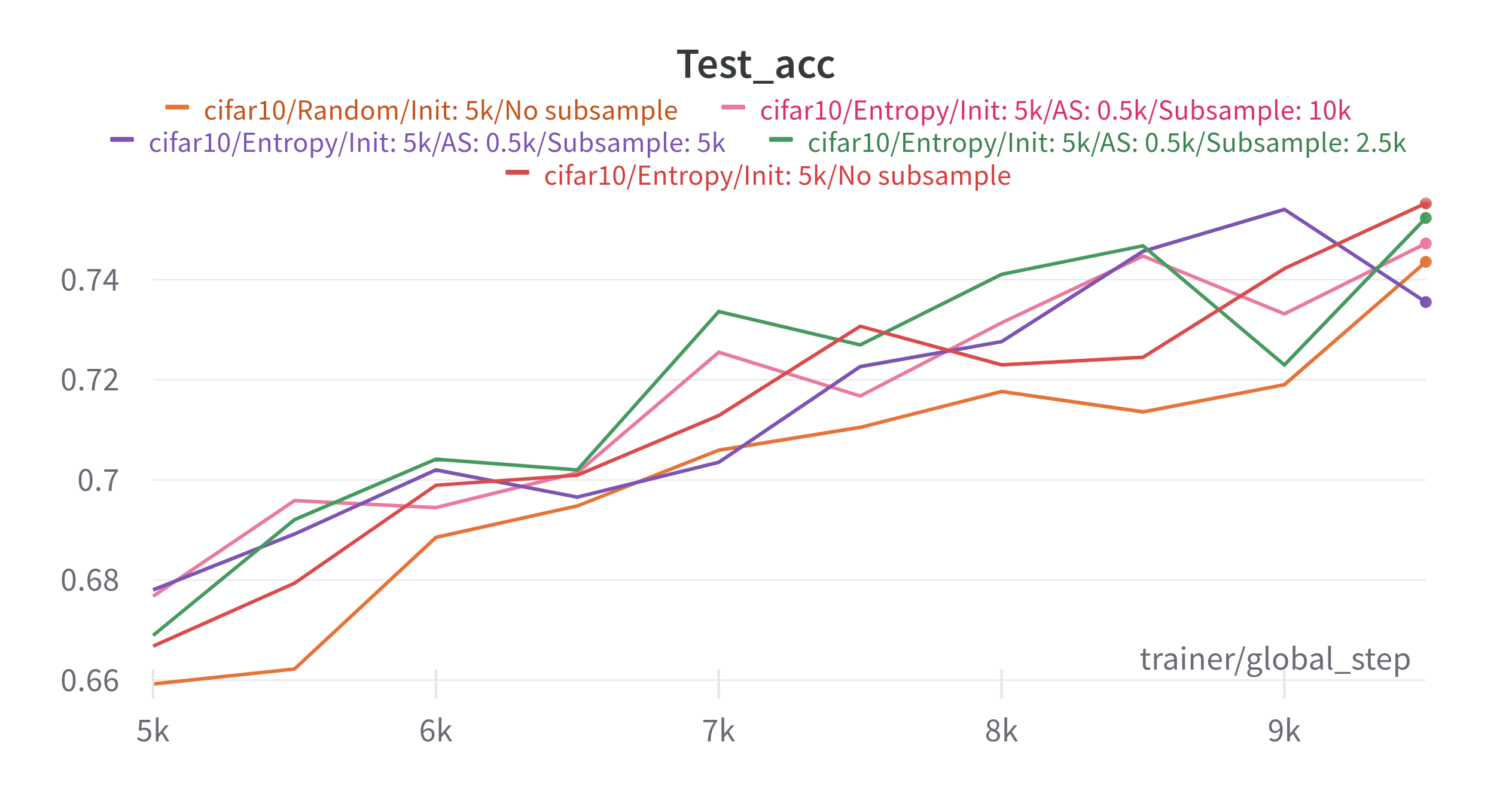}
  \caption{AF: entropy, IP: 10\%, TAS: 10\%}
  \label{fig:cifar_10_10}
\end{subfigure}
\begin{subfigure}{.5\textwidth}
  \centering
  \includegraphics[width=1.0\linewidth]{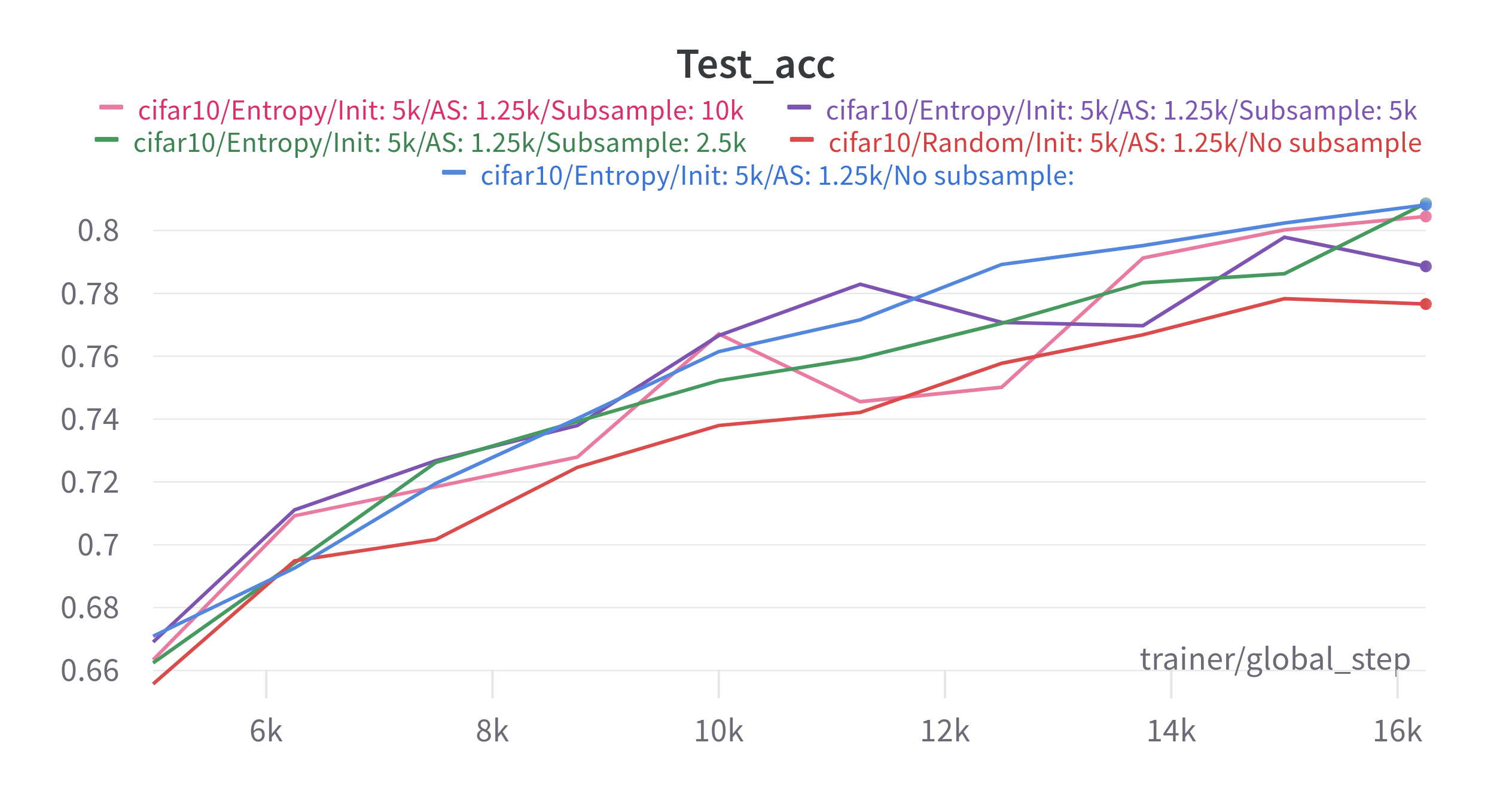}
  \caption{AF: entropy, IP: 10\%, TAS: 25\%}
  \label{fig:cifar_entr_10_25}
\end{subfigure}%
\begin{subfigure}{.5\textwidth}
  \centering
  \includegraphics[width=1.0\linewidth]{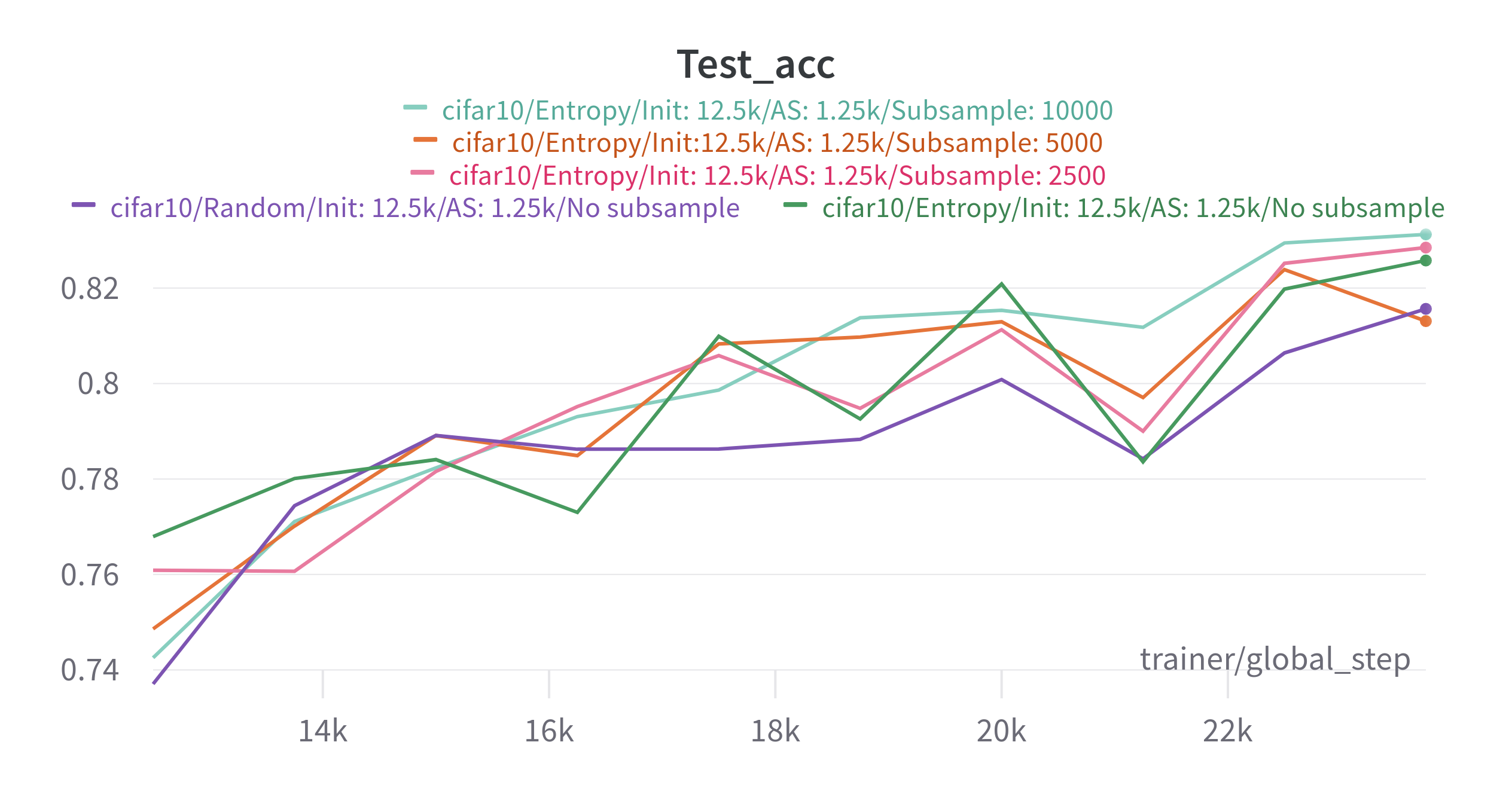}
  \caption{AF: entropy, IP: 25\%, TAS: 25\%}
  \label{fig:cifar_entr_25_25}
\end{subfigure}
\caption{Test results (classification accuracy) on CIFAR dataset with entropy acquisition function (AF), various initial pool sizes (IS), and total acquisition sizes (TAS).}
\label{fig:cifar_exps2}
\end{figure}

\begin{figure}[b]
\begin{subfigure}{.5\textwidth}
  \centering
  \includegraphics[width=1.0\linewidth]{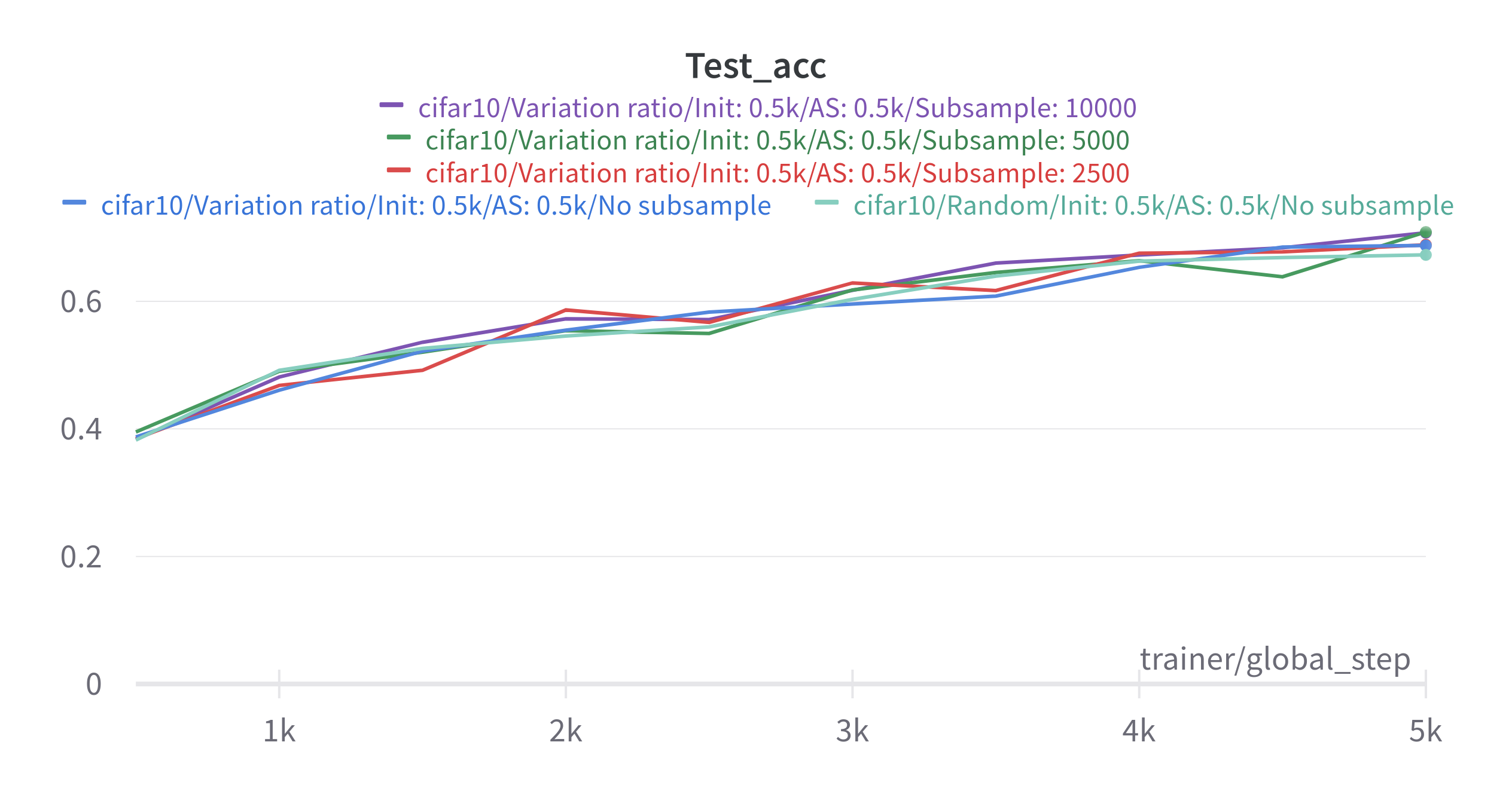}
  \caption{AF: varR, IP: 1\%, TAS: 10\%}
  \label{fig:cifar_var_1_10}
\end{subfigure}%
\begin{subfigure}{.5\textwidth}
  \centering
  \includegraphics[width=1.0\linewidth]{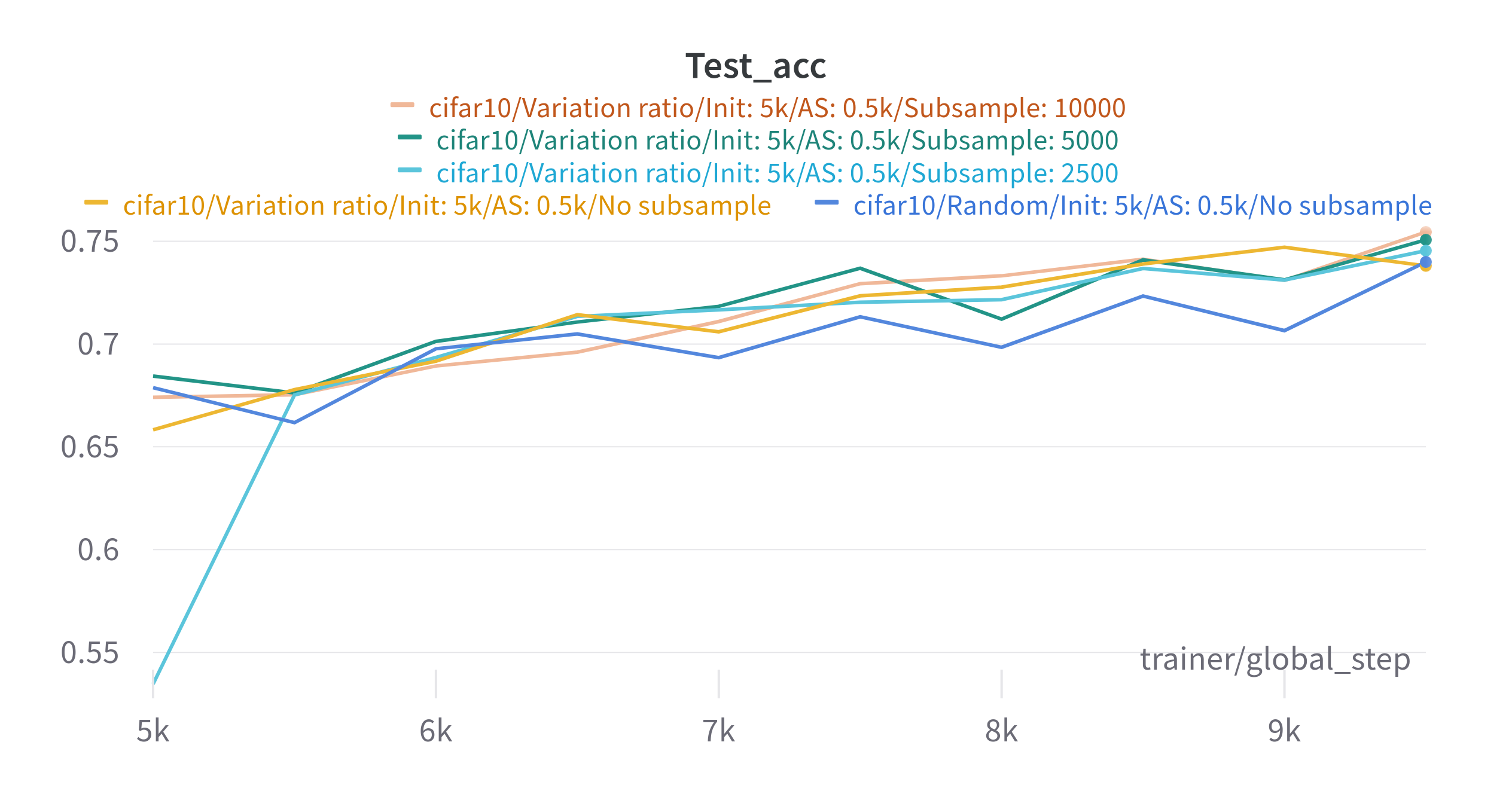}
  \caption{AF: varR, IP: 10\%, TAS: 10\%}
  \label{fig:cifar_var_10_10}
\end{subfigure}
\begin{subfigure}{.5\textwidth}
  \centering
  \includegraphics[width=1.0\linewidth]{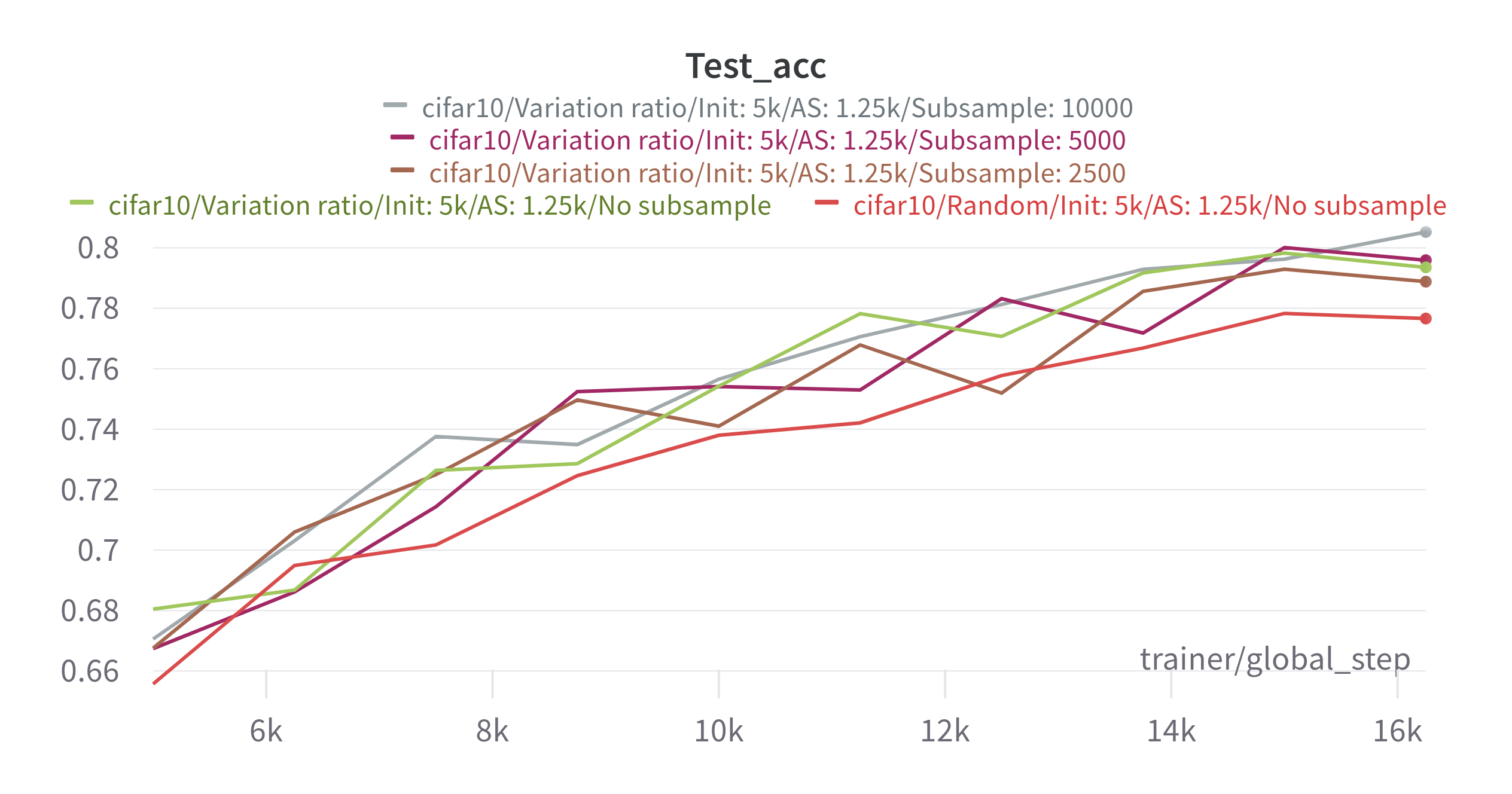}
  \caption{AF: varR, IP: 10\%, TAS: 25\%}
  \label{fig:cifar_var_10_25}
\end{subfigure}%
\begin{subfigure}{.5\textwidth}
  \centering
  \includegraphics[width=1.0\linewidth]{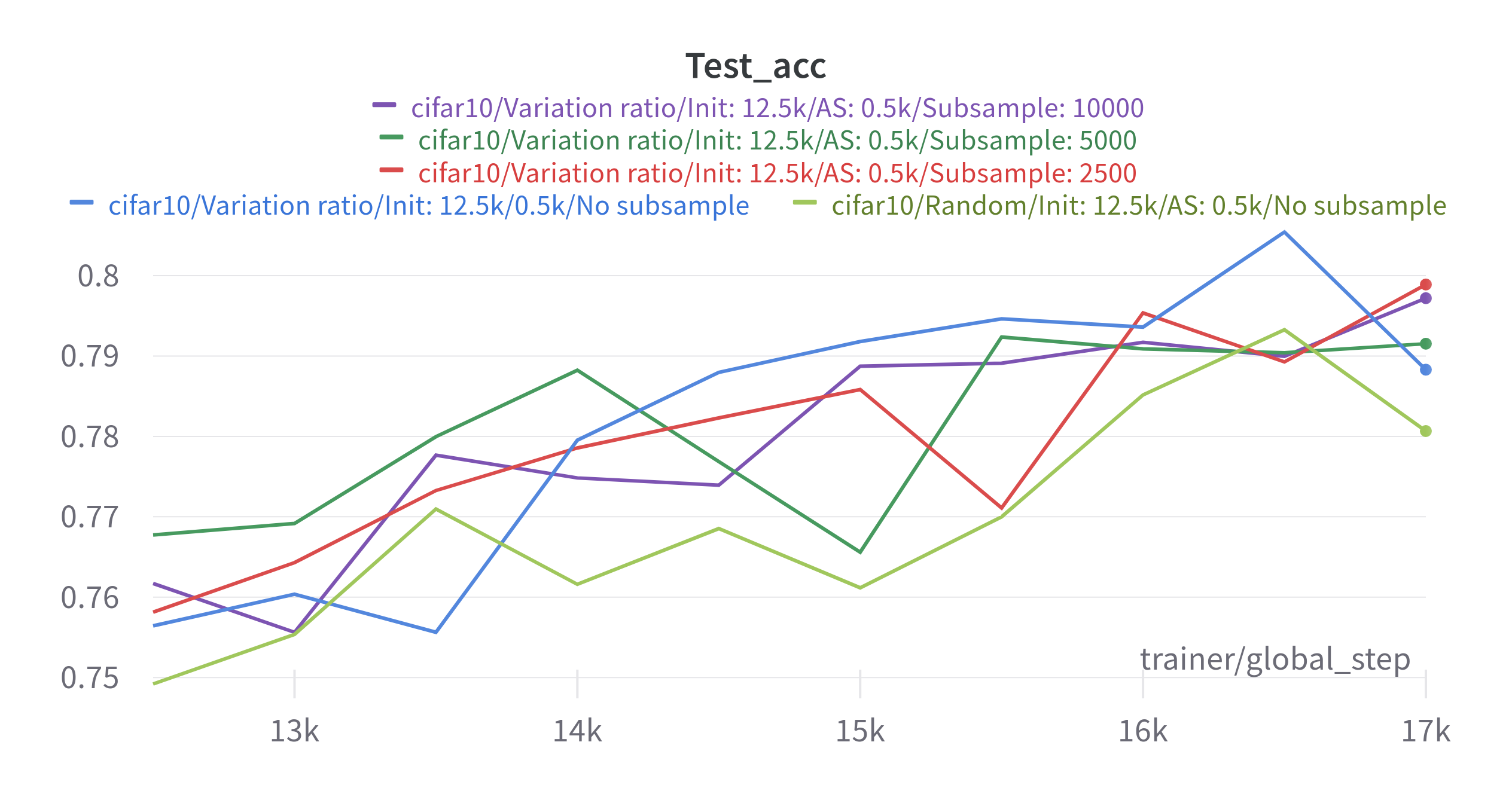}
  \caption{AF: varR, IP: 25\%, TAS: 10\%}
  \label{fig:cifar_var_25_10}
\end{subfigure}
\caption{Test results (classification accuracy) on CIFAR-10 dataset with variation ratios acquisition function (AF), various initial pool sizes (IS), and total acquisition sizes (TAS).}
\label{fig:cifar_exps3}
\end{figure}

Overall, we have conducted 24 groups of experiments, resulting in over 100 separate trainings. As Figure \ref{fig:mnist_exps2}, \ref{fig:mnist_exps3}, \ref{fig:cifar_exps2}, and \ref{fig:cifar_exps3} shows, the proposed method consistently outperforms the random selection from the entire unlabeled pool. Furthermore, the other baselines sampling from the whole unlabeled pool based on entropy or variational ratios are also outperformed by our method 14 out of 16 times despite the subsampling solution requiring significantly less computation resources and time. These additional experiments support our hypotheses that historical values of the acquisition function are good predictors of their future values and can be used for compute-efficient active learning.

\section{An example realization of the proposed method}
Let's assume our method will be executed using the CIFAR-10 dataset. Given a labeled initial pool (e.g., 10\% of the dataset) and an unlabeled pool of data points, we train a neural network and then evaluate the acquisition function on all unlabeled samples. Then, we choose the best B number (e.g., 500) of data points to label, add them to the labeled pool, and retrain the network. Deep AL is done by iterating through these steps repeatedly. 

In our solution, after the first training, evaluation, labeling, and retraining steps, we save the output of the acquisition function for each unlabeled data point and subsample the dataset based on these values into a so-called candidate pool which is small (e.g., 5\%) in comparison to the unlabeled pool. We sample in such a way that the more confident the network is on a given sample, the less likely it is to be in the candidate pool. After the second training, we evaluate the acquisition function only on the samples that are in the candidate pool and update the values of the acquisition function for these samples only. Afterward, we use the partially updated values of the acquisition function to resample the candidate pool from the unlabeled pool. The iteration continues until the iteration number (e.g., 10) is reached. 

Since we choose the size of the candidate pool, we can control the time we use for evaluating the acquisition function in every step in a non-naive way. The sampling effect not only saves us time but also compensates for the cases where suboptimal points could be chosen due to sampling. Due to sampling, the probability of highly redundant batches or unbalanced label space is much lower. 

 %%%% CIFAR-10 figure
 
\end{document}